\documentclass[10pt,twocolumn,letterpaper]{article}
\usepackage[pagenumbers]{cvpr} 

\usepackage{xspace}
\usepackage{booktabs}
\usepackage{multirow}
\usepackage{amsmath,amssymb}
\usepackage{graphicx}
\usepackage[table]{xcolor}
\usepackage{xcolor}
\newcommand{\method}{FlowScope\xspace}
\newcommand{\pending}{\textcolor{orange}{pending}}   
\usepackage{pgfplots}
\pgfplotsset{compat=1.17}
\usepgfplotslibrary{groupplots}
\usetikzlibrary{calc}

\newcommand{\note}[1]{\textcolor{blue}{[#1]}}                      
\newcommand{\figplaceholder}[2]{
  \fbox{\parbox[c][#1][c]{0.95\linewidth}{\centering\small\textcolor{gray}{#2}}}}

\definecolor{baseD}{HTML}{5E5E5E}
\definecolor{baseM}{HTML}{A3A3A3}
\definecolor{baseL}{HTML}{DEDEDE}
\definecolor{oursD}{HTML}{1F5FA8}
\definecolor{oursM}{HTML}{6A9FD8}
\definecolor{oursL}{HTML}{C9DDF3}

\definecolor{effScratch}{HTML}{8A9099}
\definecolor{effKin}{HTML}{2F6DB5}
\definecolor{effFlow}{HTML}{E07A2F}
\definecolor{effMuted}{HTML}{5B6472}
\definecolor{effGrid}{HTML}{E3E6EA}

\definecolor{encOTS}{HTML}{B9BEC6}
\definecolor{encIH}{HTML}{1F9E89}
\definecolor{encMuted}{HTML}{5B6472}
\definecolor{encGrid}{HTML}{E3E6EA}

\newcommand{\para}[1]{\noindent\textbf{#1}\enspace}

\AtEndPreamble{%
  \crefname{appendix}{Supp.}{Supp.}%
  \Crefname{appendix}{Supp.}{Supp.}%
}
\definecolor{cvprblue}{rgb}{0.21,0.49,0.74}
\usepackage[pagebackref,breaklinks,colorlinks,allcolors=cvprblue]{hyperref}

\def\confName{CVPR}
\def\confYear{2027}

\title{\method: 3D Scene Flow Pretraining Improves Surgical Robot Policies}

\author{
John J. Han$^{1}$ \quad
Adam Schmidt$^{2}$ \quad
Jingpei Lu$^{2}$ \quad
Zooey He$^{2}$ \quad 
Muhammad Abdullah Jamal$^{2}$ \quad \\[2pt]
Jie Ying Wu$^{1}$ \quad 
Omid Mohareri$^{2}$ \\[2pt]
$^{1}$Department of Computer Science, Vanderbilt University \qquad \\[2pt]
$^{2}$Intuitive Surgical Inc. \\[2pt]
}

\begin{document}
\maketitle
\begin{abstract}
Imitation learning for surgical robots is limited by the cost of collecting demonstrations.
Meanwhile, many robot-assisted procedures are recorded as stereo video, yet policies cannot learn from them directly; clinical video carries no action labels. 
We present \method, which pretrains a surgical policy network on unlabeled clinical video by forecasting scene flow, the metric 3D motion of scene points, with labels computed by off-the-shelf stereo depth and optical flow models.
We pretrain the trunk of an Action Chunking Transformer (ACT) on a dataset of 1{,}149 clinical procedures (1.35M clips) and discard the flow head afterward, so the deployed policy is a standard ACT with no added inference cost.
On a real surgical robotic system, \method raises needle pickup success from 54\% to 90\% and first-handover success on needle regrasp from 4\% to 52\% (full task: 2\% to 14\%) over training from scratch ($n=50$ trials each), and retains its gains with a generative flow-matching action head.
Ablations indicate that metric 3D motion is the strongest pretraining signal among those tested; kinematics, 2D tracks, depth alone, and scale-free 3D flow, each paired with kinematics, yield smaller or no gains.
Representation analyses suggest that scene-flow pretraining encodes metric needle depth more accurately, that kinematics pretraining increases reliance on the robot state, and that needle-depth probes follow the ordering of real-robot success where offline validation error does not.
\end{abstract}
\section{Introduction}
\label{sec:intro}
Robot-assisted surgery is a promising but underexplored frontier for robot learning.
Automating even routine suturing subtasks, such as needle pickup or needle handover, could reduce surgeon fatigue and ultimately improve patient outcomes.
Imitation learning has recently made dexterous behaviors like these tractable in general-purpose manipulation~\cite{zhao2023act, chi2023diffusion, black2024pi0}, following a simple recipe: collect more demonstrations, train better policies.
In surgical robotics, the first half of this recipe is the bottleneck. 

Surgical demonstrations are scarce and expensive.
Each one requires an expert operator, often an experienced robotic surgeon, a carefully prepared physical setup, and time on a costly system.
Simulation offers a potential alternative, but accurately modeling deformable tissues and realistic endoscopic scenes remains challenging, limiting reliable transfer of learned policies to physical surgical robots~\cite{han2026endopbr,liang2024real,scheikl2022sim}.

Pretraining offers another way to reduce the demonstration burden.
It has transformed language~\cite{radford2018improving, radford2019language, devlin2019bert} and vision~\cite{he2022masked, oquab2024dinov2, simeoni2025dinov3}. Robotics has followed, from the RT series~\cite{brohan2022rt, brohan2023rt, o2024open} to vision-language-action models built on web-scale data~\cite{black2024pi0, intelligence2025pi05}.
The natural next step is to fine-tune these generalist models on surgical tasks.
Yet vision models pretrained on general-domain images transfer poorly to surgical scenes~\cite{han2026dart, perez2026surglavi, de2025scaling}, and the robot data behind generalist policies looks nothing like the inside of a patient.
Indeed, in our experiments, a fine-tuned $\pi_{0.5}$ trails \method on both surgical tasks (Sec.~\ref{sec:exp}).
Surgical policies need surgical pretraining data.

Such data already exists at scale.
Many robot-assisted procedures are recorded through an endoscope, producing a large and growing corpus of video from the domain a policy will be deployed in.
This video, however, cannot serve as demonstrations directly.
Imitation learning needs action labels, but clinical video carries none; the kinematics of clinical systems are rarely available outside the manufacturer.
Even when they are available, they are offset from where the instruments appear in the image, due to errors in the forward kinematics and calibration~\cite{ye2016real,kim2024srt}. 
Actions recovered from them do not match what the policy sees.

What, then, should a surgical policy learn from unlabeled stereo video?
Our answer is \emph{motion}.
What a surgeon does is visible in how the scene moves: the instruments sweep toward the needle, the needle lifts from the tissue, and the tissue deforms under traction.
Because the endoscope is stereo, this motion can be estimated in metric 3D with off-the-shelf stereo depth and optical flow models.
The resulting scene flow is an action-\textit{like} signal that is read directly from the pixels, needs no robot labels, and does not inherit the errors of the kinematics.

\begin{figure*}[t]
  \centering
  \includegraphics[width=\linewidth]{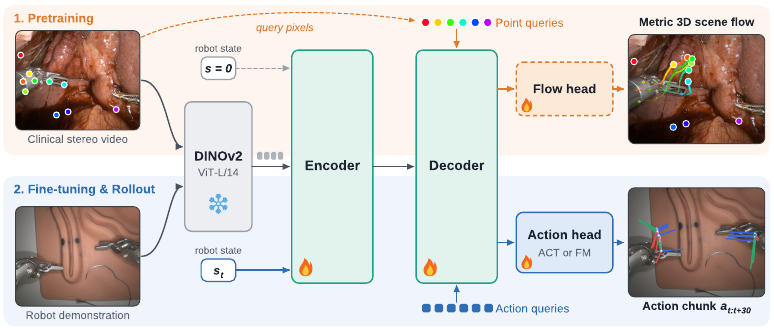}
  \caption{\textbf{\method.} Each image is encoded by a frozen DINOv2 pretrained on surgical images. \emph{Pretraining} (top): point queries read the image features at their pixels, and the flow head predicts each point's metric 3D position and future trajectory, supervised with an L1 loss on scene flow pseudo-labeled by off-the-shelf stereo and optical-flow models. The robot state is set to zero. \emph{Fine-tuning} (bottom): the point queries and flow head are discarded, and the pretrained trunk is trained on robot demonstrations to predict action chunks with either the ACT or a flow-matching (FM) action head, giving a standard ACT policy. The CVAE (encoder not shown) is used only during training; z = 0 at test, similar to standard ACT rollout.}
  \label{fig:arch}
\end{figure*}

We present \method, a pretraining recipe for surgical policies built on scene flow.
Using off-the-shelf stereo depth~\cite{wen2026fast} and optical flow~\cite{wang2024seaflow} models, we convert raw surgical video into metric 3D trajectories of points on tissue, needles, and instruments.
We then pretrain the trunk of an Action Chunking Transformer (ACT) model to take the endoscope image as input and predict the 3D positions and future trajectories of queried points.
Pretraining adds only a set of point queries and a lightweight flow head, both discarded afterward, so the fine-tuned model is a standard ACT policy and \method requires no changes at deployment.

We pretrain \method with 1{,}149 clinical procedures, about 1.35M training clips, and fine-tune it on two subtasks on a minimally invasive surgical robotic system: needle pickup and needle regrasp.
On both tasks, \method substantially improves real-robot success over training from scratch: from 54\% to 90\% on pickup, and from 4\% to 52\% on the first handover of regrasp.
It also outperforms Diffusion Policy~\cite{chi2023diffusion} and the finetuned vision-language-action model $\pi_{0.5}$~\cite{intelligence2025pi05}.
The gains are not tied to ACT's regression head; they persist when we replace it with a generative flow-matching head.
Pretraining also improves data efficiency: with half of the demonstrations, \method matches a policy trained from scratch on all of them on needle pickup (52\% vs.\ 54\%) and surpasses it on needle regrasp (32\% vs.\ 4\% first-handover success).

Two findings stand out.
First, scene flow alone is a better pretraining signal than robot kinematics for surgical robot learning.
We expected the kinematics recorded alongside the video to be useful supervision, since they describe exactly what a policy must output.
Yet pretraining on kinematics gives little to no gain over training from scratch, and sometimes hurts; adding kinematics to scene-flow pretraining lowers success.
Second, metric 3D scene flow outperforms its alternatives based on 2D point tracks, depth-only supervision, and scale-free 3D flow. Metric scale, which the stereo endoscope provides, is what makes \method work.


To understand why, we look inside the policies.
Kinematics-pretrained policies devote nearly three times as much visual attention to the robot-state input as other policies, and keep doing so after fine-tuning.
Flow-pretrained policies instead encode the needle's metric depth more accurately than any other initialization.
\method teaches the policy to locate what matters in 3D from visual input only.

In summary, our contributions are:
\begin{itemize}
    \item \method, a recipe that pretrains surgical policy networks to forecast metric 3D scene flow from unlabeled clinical stereo video, requiring no robot labels and no changes to the deployed policy.
    \item A real-robot evaluation on two surgical tasks, showing that \method improves success and data efficiency over training from scratch, kinematics pretraining, and strong baselines, with both regression and generative heads.
    \item Controlled ablations showing that the benefit comes from \emph{metric 3D} motion: 2D tracks, depth alone, and scale-free 3D flow give smaller or no gains when added to kinematics, and kinematics alone are a weaker signal than scene flow.
    \item Representation analyses showing that flow pretraining improves the policy's encoding of metric needle depth, that kinematics pretraining shifts attention toward the robot state, and that such probes, unlike offline validation error, track real-robot success. 
\end{itemize}
\section{Related Work}
\label{sec:related}
\para{Visuomotor imitation learning.}
Action chunking with transformers (ACT)~\cite{zhao2023act} and Diffusion Policy~\cite{chi2023diffusion} made imitation learning practical for dexterous manipulation.
Vision-language-action models~\cite{black2024pi0,intelligence2025pi05} go further: they pretrain on large robot and web datasets and adapt to a new task with a few hours of demonstrations.
Most of these policies build on a pretrained visual backbone, and frozen self-supervised encoders such as DINOv2~\cite{oquab2024dinov2} are a common choice~\cite{abc2026}.
\method keeps this design and changes only what the policy is pretrained on.

\para{Learning policies from video motion.}
Point tracks and flow are a popular interface between action-free video and control.
ATM~\cite{wen2024atm} predicts 2D tracks of arbitrary points and conditions a policy on them; TraceGen~\cite{lee2026tracegen}, 3PoinTr~\cite{hung20263pointr}, and PointZero~\cite{duisterhof2026pointzero} predict object or scene trajectories; and ARM4R~\cite{niu2025arm4r} and SPA~\cite{zhu2025spa} pretrain 3D- and 4D-aware representations for embodied learning.
In surgery, SurgFlow~\cite{chen2026surgflow} learns from stereo video of teleoperated dVRK demonstrations of each task.
It segments the target object, predicts the object's future 3D flow and where the tool should make contact, and converts these into actions with an optimization-based controller and contact-gated grasping.
\method differs from these methods in three ways.
It pretrains on \emph{clinical} video from many procedures, rather than on video of the target task.
It predicts metric 3D flow over the whole scene, rather than 2D tracks, unscaled flow from monocular video, or segmented objects, so it needs no object masks.
And flow is only a pretraining target: the deployed policy is a standard learned policy that maps images to actions, with no flow prediction or controller at test time.

\para{What makes visual representations useful for control?}
Burns \etal~\cite{burns2023makes} found that how well a ViT's attention segments objects predicts out-of-distribution manipulation success better than generic downstream accuracy.
Wen \etal~\cite{wen2025datacentric} trace the strengths and failures of pretrained vision models for robot learning to whether they form object-centric representations.
We follow this line of analysis but apply it to the policy itself.
Instead of probing the vision encoder, we probe the pretrained and fine-tuned policy trunks for the needle, the instruments, and their metric depth, and relate the probes to closed-loop success.

\para{Surgical robot learning.}
Research platforms such as the dVRK~\cite{kazanzides2014dvrk} enabled imitation learning for surgical subtasks.
The Surgical Robot Transformer~\cite{kim2024srt} showed that relative action parameterizations mitigate the kinematic error of cable-driven arms.
Clinical recordings, in turn, have been used widely for surgical perception, including phase recognition, segmentation, and depth estimation~\cite{han2026dart}.
Recent work has begun to bring this video into policy learning, by turning it into actions.
SurgVIL~\cite{chen2026surgvil} estimates tool kinematics from monocular clinical video and co-trains policies on these pseudo-actions together with robot demonstrations, evaluating needle pickup on a real robot.
Cosmos-H-Surgical~\cite{he2025cosmos} fine-tunes a video world model on annotated surgical clips, labels its generated videos with an inverse dynamics model, and uses them as extra training data, but reports offline trajectory error rather than real-robot rollouts.
Surgical WAM~\cite{bao2026surgical} pretrains a world-action model on action-free video and evaluates it only in simulation.
\method differs in what the policy learns from video.
Instead of actions, estimated or generated, it predicts the metric 3D motion of the whole scene, which stereo video measures directly and which needs neither action labels nor video generation.
Furthermore, our experiments show that pretraining on the recorded kinematics, which are at least as accurate as kinematics estimated from video, gives no consistent gain in our real-robot evaluations, while leading the policy to rely on the robot state rather than the visual observations (\cref{sec:analysis}).
\section{Method}
\label{sec:method}

\method learns a visuomotor policy in two stages, shown in \cref{fig:arch}.
First, we pretrain the policy's transformer trunk on clinical surgical video, using only a pretext task that needs no robot actions: predicting where points in the scene are in 3D and how they will move.
Second, we fine-tune the pretrained trunk on a small set of robot demonstrations, either with the standard ACT recipe or with a generative flow-matching action head.

\subsection{Problem Setup}
\label{sec:problemsetup}
A policy $\pi_\theta(a_{t:t+H} \mid o_t, s_t)$ maps the left endoscope image $o_t$ and the robot state $s_t$ to a chunk of $H$ future actions.
Unlike tabletop setups, which often add wrist cameras, minimally invasive surgery offers a single viewpoint: the endoscope.
We use only its left image, so stereo is needed only to construct pretraining targets and the deployed policy remains single-view; feeding both views to the policy did not improve success and doubles the cost (\cref{sec:supp_exp_stereo}).
The standard approach trains $\pi_\theta$ from scratch on a dataset $\mathcal{D}$ of teleoperated demonstrations of the target task, such as needle pickup, which are expensive to collect.
We additionally assume a much larger corpus $\mathcal{V}$ of clinical stereo videos.
Unlike $\mathcal{D}$, these videos carry no action labels, but they show the same kind of scene and often the same skills, such as grasping and passing needles.
Our goal is to use $\mathcal{V}$ to pretrain the trunk of $\pi_\theta$ so that imitation learning on $\mathcal{D}$ becomes more effective.
As the pretraining target we use metric 3D scene flow: for a queried point in the current frame, predict its 3D position and its 3D displacement over time.
This target is computed from video alone, so pretraining uses neither the robot state nor any actions.

\subsection{3D Motion Targets from Clinical Video}
\label{sec:data}

\begin{figure}[t]
  \centering
  \includegraphics[width=\linewidth]{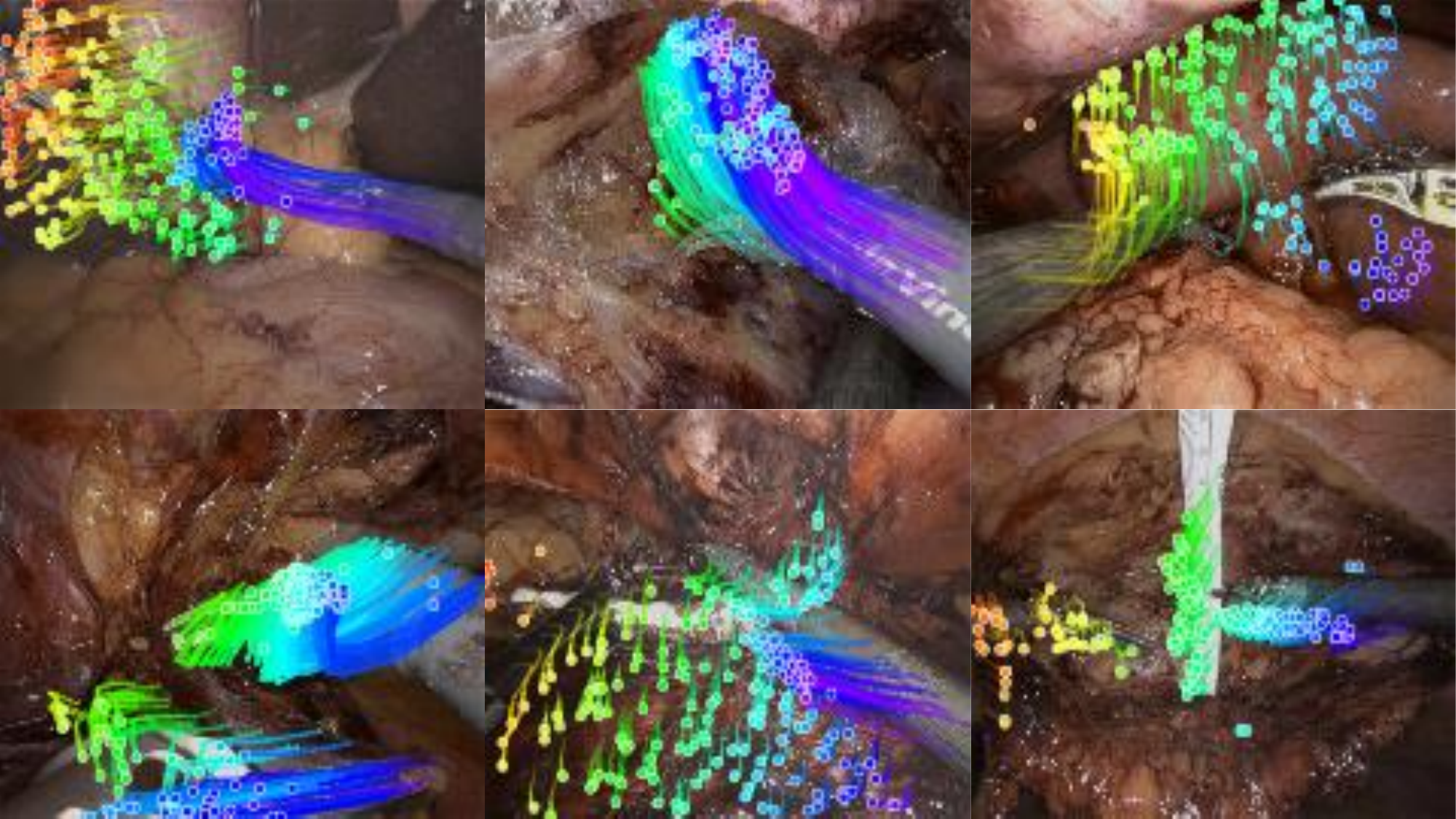}
  \caption{\textbf{Scene-flow pseudo-labels on clinical video.} Each track shows where a point on the anchor frame moves over the next 10 keyframes (1.7\,s), computed from stereo depth and optical flow and kept only where both pass consistency checks. More visualizations can be seen in \Cref{fig:supp_flow_examples}.}
  \label{fig:flow_vis}
\end{figure}

\para{Pretraining corpus.}
We use stereo endoscope recordings of clinical procedures performed with a minimally invasive surgical robotic system: 1{,}175 procedures, split by video into 1{,}149 training and 26 validation videos (\cref{fig:flow_vis}). 
All videos were thoroughly de-identified to ensure patient privacy, and data was collected with consent from anonymized surgical procedures, under our IRB-equivalent institutional review process. 
From each video we take candidate \emph{anchor frames} every 10 frames, each paired with the scene's motion over the following 100 frames (1.7\,s).
We keep an anchor only if the endoscope stays still over this window (within 0.5\,mm and 0.5$^\circ$), so that all labeled motion comes from tissue and instruments rather than the camera, and if the surgeon drives at least one instrument for the whole window.
We read both conditions from the logged kinematics, which serve only this selection step: they are never an input or a target during pretraining.
After filtering, the corpus yields 1.35M anchors.

\para{Metric depth and optical flow.}
We estimate dense depth from disparity for every stereo pair with Fast-FoundationStereo~\cite{wen2026fast} and dense optical flow between consecutive keyframes with SEA-RAFT~\cite{wang2024seaflow}; both generalize well to surgical images~\cite{han2026scared,schmidt2025point}.
Pseudo-labels from both models can be wrong at occlusion boundaries, on specular highlights, and on textureless tissue, so we use a prediction only if it is self-consistent: disparity estimated from the left and from the right image must agree within 1\,px, and flow followed forward and then backward must return within 1\,px (plus 5\% of its magnitude).
These checks remove regions seen by only one eye and points that become occluded, for example when an instrument passes over tissue.
~\cref{sec:supp_filters} lists all filters with their thresholds.

\para{3D point trajectories.}
At each anchor frame we track a grid of pixels through the next $K{=}10$ keyframes by chaining the keyframe-to-keyframe flow, and lift every tracked pixel to a 3D point in the camera frame using the depth at that keyframe.
This gives each point a start position $P_0 \in \mathbb{R}^3$ and displacements $\Delta P_k = P_k - P_0$, $k = 1, \dots, K$, in millimeters.
A point stops being tracked at the first keyframe where it fails a check, so each (point, keyframe) pair carries its own validity flag.
Since most of a surgical scene is static in any given second, uniform sampling would mostly teach the model that nothing moves.
We therefore store 512 points per anchor, 256 sampled uniformly and 256 with probability proportional to how far they move, and at training time use 256 of them as queries, up to half drawn from points that move more than 5\,mm. 

\subsection{Model and Pretraining}
\label{sec:model}

\para{Overview.}
We build on ACT~\cite{zhao2023act}, a DETR-style~\cite{carion2020end} encoder-decoder transformer widely used for imitation learning.
We keep ACT's architecture unchanged so that the pretrained weights transfer directly, and change only what it is trained to predict.
During pretraining, the network receives a single image and a set of point queries, and predicts the metric 3D position and future 3D trajectory of each queried point.
Everything that depends on robot data is inactive: the robot-state token and the CVAE latent token are set to zero, and ACT's action queries and action head receive no loss.
We keep these parts only so that the architecture matches ACT exactly at fine-tuning time.

\para{Image tokenizer.}
Surgical vision foundation models outperform general-purpose ones on surgical tasks~\cite{han2026dart}.
We therefore train a DINOv2 ViT-L/14~\cite{oquab2024dinov2} on surgical video (details and comparison studies in~\cref{sec:supp_encoder}).
It stays frozen in every stage of \method and acts as a tokenizer, mapping the left image at $518{\times}518$ to $37{\times}37 = 1{,}369$ patch tokens, which the transformer encoder processes together with the inactive state and latent tokens.

\para{Point queries.}
The decoder receives one query per tracked point.
For a point at image position $u$, the query is
\begin{equation}
  q(u) = W_u\,\gamma(u) + W_f\,\phi(u) + e_{\text{pt}},
\end{equation}
where $\gamma(u)$ is a Fourier encoding of the position, $\phi(u)$ is the encoder output bilinearly sampled at $u$, and $e_{\text{pt}}$ is a learned embedding marking the query as a point rather than action.
Each query thus starts from what the ACT encoder sees at that pixel, and the decoder refines it by cross-attending to the whole image.

\para{Flow head and loss.}
A shallow MLP maps the decoder output of each point query to the predicted 3D start position $\hat P_0$ and displacements $\Delta\hat P_1, \dots, \Delta\hat P_K$.
Let $Q$ be the number of point queries in a sample and $\Omega$ the set of valid (point, keyframe) pairs.
We train with an L1 loss on the displacements and on the start positions:
\begin{align}
  \mathcal{L}_{\text{disp}} &= \frac{1}{|\Omega|} \sum_{(i,k) \in \Omega}
      \big\lVert \Delta\hat P_{i,k} - \Delta P_{i,k} \big\rVert_1, \\
  \mathcal{L}_{\text{pos}} &= \frac{1}{Q} \sum_{i=1}^{Q}
      \big\lVert \hat P_{i,0} - P_{i,0} \big\rVert_1, \\
  \mathcal{L}_{\text{flow}} &= \mathcal{L}_{\text{disp}} +\, \mathcal{L}_{\text{pos}}.
\end{align}
All targets are normalized with statistics computed on the training videos.
$\mathcal{L}_{\text{pos}}$ asks the model where each point is in metric 3D, and $\mathcal{L}_{\text{disp}}$ asks where it will go.
Since future motion is often ambiguous from a single frame, we also test a generative flow-matching~\cite{lipman2023flow} flow head in place of the L1 head (\cref{sec:ablation}).

\subsection{Fine-tuning and Deployment}
\label{sec:finetune}

\para{From pretraining to a policy.}
We initialize the policy from the pretrained checkpoint with the lowest validation flow error.
For fine-tuning, we remove the point queries and the flow head, and switch the robot-specific parts of ACT back on: the robot state enters through the state token, and the CVAE encoder produces the latent from the action chunk.
These parts and the action head are randomly initialized, exactly as in an ACT trained from scratch; the encoder, the decoder, and the input projections start from pretraining.
We then train the whole policy on $\mathcal{D}$ with ACT's standard $\mathcal{L}_1$ and KL objective, keeping the image tokenizer frozen.

\para{Action representation.}
Following prior work~\cite{kim2024srt}, we express all kinematics in the camera frame and act in end-effector Cartesian space.
For each arm, an action is a translation relative to the previous step, an absolute rotation (a $3{\times}3$ rotation matrix), and an absolute gripper angle: 13 dimensions per arm, 26 for both.
The policy predicts a chunk of $H{=}30$ actions covering 2\,s.

\para{Generative action head.}
\label{sec:genhead}
To test whether the pretrained trunk also helps a generative policy, we replace ACT's CVAE and action head with a flow-matching action head on the same trunk, in the spirit of Diffusion Policy~\cite{chi2023diffusion} and the action expert of $\pi_0$~\cite{black2024pi0}.
The head is a 4-block transformer with adaptive layer normalization~\cite{peebles2023dit} that denoises the whole 30-step chunk, conditioned on the decoder's output for each step, and samples a chunk with 10 Euler steps.
It is trained from scratch during fine-tuning on top of any pretrained trunk (\cref{sec:supp_fm}).

\para{Deployment.}
At test time the policy is a standard ACT (or ACT with the flow-matching head): it receives only the left endoscope image and the robot state, and needs no depth, flow, or other pseudo-labels.
\section{Experiments}
\label{sec:exp}

We evaluate \method on a real surgical robotic system and ask three questions.
Does pretraining on clinical video improve real-robot success, with either action head (\cref{sec:main})?
Does it reduce the number of demonstrations needed (\cref{sec:efficiency})?
And which part of the pretraining target is responsible for the gain (\cref{sec:ablation})?

\subsection{Experimental setup}
\label{sec:setup}


\para{Tasks.}
We evaluate two suturing subtasks of increasing difficulty.
In \emph{needle pickup}, the dominant (right) instrument grasps a suture needle lying on the pad and lifts it.
\emph{Needle regrasp} starts the same way, but the needle is then handed to the non-dominant instrument and back to the dominant one.
Surgeons perform this regrasp to bring the needle into a good pose for driving it through tissue, which is the next step of suturing.
Regrasp therefore requires precise coordination of both arms over a long sequence, and we score it in three stages: pickup, the first handover, and the second handover, which completes the task.

\para{Robot and data.}
All experiments run on a minimally invasive surgical robotic system, with a suture pad placed inside a mock abdominal cavity (\cref{fig:supp_setup}).
We collected 898 teleoperated demonstrations of needle pickup and 2{,}255 of needle regrasp, including DAgger~\cite{ross2011reduction} data. Additional details are in the supplementary material (\cref{sec:supp_data}).
Every policy, baselines included, is fine-tuned on the same demonstrations and evaluated on the same physical setup.

\para{Evaluation protocol.}
Real-robot success varies across checkpoints and validation loss does not predict rollout performance, so we evaluate each policy at five fine-tuning checkpoints (every 20k steps up to 100k) on the same 10 predefined needle configurations (\cref{fig:supp_configs}), with 50 trials per policy.
For pickup, the policy may make up to three attempts per trial, and we report success on the first attempt and within three attempts.
For regrasp, the policy gets a single attempt, and we report the fraction of trials that complete each stage, together with the mean number of stages completed (from 0 to 3).
Additional details are in \cref{sec:supp_robot}.

\para{Implementation.}
Pretraining runs for 100k steps on 4 H200 GPUs with 128 samples per GPU.
Fine-tuning runs for 100k steps on a single 40\,GB A100 with a batch size of 128, photometric augmentation, and noise added to the robot state (1.5\,mm and 3$^\circ$) to improve robustness against noisy measurements. Both stages use a base learning rate of $10^{-5}$, scaled with the square root of the batch size, and the frozen surgical DINOv2 tokenizer at 518\,px.
Policies with the flow-matching action head additionally use a cosine learning-rate schedule, an exponential moving average of the weights, and full precision in the head, which we found necessary for stable training (see \cref{sec:supp_fm_v2}).

\para{Baselines.}
We compare against the following baselines.
\begin{itemize}
  \item \textbf{No pretraining.} The same policy (ACT or the flow-matching action head) trained from scratch on the demonstrations.
  \item \textbf{External policies.} Diffusion Policy~\cite{chi2023diffusion} and the vision-language-action model $\pi_{0.5}$~\cite{intelligence2025pi05}. For a fair comparison, Diffusion Policy uses the same frozen surgical DINOv2 as \method; everything else follows its standard real-robot recipe. We fine-tune $\pi_{0.5}$ both fully and with LoRA~\cite{hu2021lora}. See \cref{sec:supp_baselines} for more details.
  \item \textbf{Kinematics pretraining.} The natural alternative way to use the same clinical videos: train a vanilla ACT to predict the surgeon's recorded kinematics, exactly as in behavior cloning, with the recorded robot state as input and the same action format as the demonstrations. It uses the same data and training budget as \method. \cref{sec:ablation} also tests predicting kinematics and scene flow together.
\end{itemize}
\subsection{Does pretraining on clinical video help?}
\label{sec:main}

\begin{table}[t]
\centering
\small
\setlength{\tabcolsep}{4pt}
\begin{tabular}{@{}llcccccc@{}}
\toprule
 & & \multicolumn{2}{c}{Pickup (\%)} & \multicolumn{4}{c}{Regrasp} \\
\cmidrule(lr){3-4}\cmidrule(l){5-8}
Pretraining & Policy & 1st & $\le$3 & S1 & S2 & S3 & Stages \\
\midrule
\multicolumn{8}{@{}l}{\textit{External baselines}} \\
---  & Diffusion Policy        & 28 & 62 & 42 & 18 & 2 & 0.62 \\
VLA  & $\pi_{0.5}$ (full FT)   & 42 & 42 & 24 & 8  & 0 & 0.32 \\
VLA  & $\pi_{0.5}$ (LoRA)      & 38 & 38 & 20 & 10 & 0 & 0.30 \\
\midrule
\multicolumn{8}{@{}l}{\textit{ACT trunk: pretraining $\times$ action head}} \\
None       & ACT & 22 & 54 & 28 & 4  & 2 & 0.34 \\
None       & FM  & 14 & 58 & 36 & 20 & 0 & 0.56 \\
Kinematics & ACT & 30 & 60 & 46 & 12 & 2 & 0.60 \\
Kinematics & FM  & 6  & 56 & 14 & 4  & 0 & 0.18 \\
\rowcolor{blue!6}
Flow (ours) & ACT & \textbf{54} & \textbf{90} & \textbf{76} & \textbf{52} & \textbf{14} & \textbf{1.42} \\
\rowcolor{blue!6}
Flow (ours) & FM  & \textbf{54} & \textbf{90} & 68 & 40 & 12 & 1.20 \\
\bottomrule
\end{tabular}
\caption{\textbf{Real-robot success (\%).} Pickup: success on the first attempt and within three attempts. Regrasp: subtasks completed (S1 pickup, S2 first handover, S3 second handover/full task) and the mean number of stages completed (max 3). $n = 50$ trials per policy (10 needle configurations $\times$ 5 checkpoints). ACT: the standard action-chunking transformer with its CVAE and L1 action loss. FM: the flow-matching action head (\cref{sec:genhead}). 95\% Wilson intervals for pickup ($\le$3): scratch [40, 67], kinematics [46, 72], \method [79, 96] with the ACT head.}
\label{tab:main_results}
\end{table}

\Cref{tab:main_results} summarizes the main results.

\para{Baselines leave a large gap.}
Trained from scratch, ACT picks up the needle within three attempts in only 54\% of trials, and Diffusion Policy in 62\%.
$\pi_{0.5}$ behaves differently: it has the highest first-attempt success among the baselines (42\% with full fine-tuning), but it never recovers once its first attempt fails, so its success stays at 42\% after three attempts.
LoRA and full fine-tuning perform about the same, even though full fine-tuning uses twice the GPUs.
On regrasp, every baseline struggles to get past the pickup stage: at most 20\% of trials reach the first handover, and none completes more than 2\% of trials.

\para{Scene-flow pretraining helps; kinematics pretraining does not.}
With the ACT head, \method raises pickup success from 54\% to 90\% and first-attempt success from 22\% to 54\%.
The gain grows with the difficulty of the task.
On regrasp, \method reaches the first handover in 52\% of trials, compared with 4\% for ACT from scratch and 12\% for kinematics pretraining, and it is the best policy at every stage, including full-task completion (14\%).
In paired comparisons on regrasp, \method outperforms ACT from scratch in 31 trials and underperforms in 4 ($p < 10^{-5}$).

Kinematics pretraining, in contrast, barely helps (60\% pickup success, against 54\% from scratch), even though its pretraining labels are actions in exactly the format of the demonstrations.
In our setting, pretraining therefore helps when the target is the scene's 3D motion, but not when it is standard behavior cloning on the surgeon's recorded motion.

\para{The gain holds for a generative action head.}
Replacing ACT's head with the flow-matching action head leaves the conclusion unchanged.
With this head, \method succeeds at pickup in 90\% of trials, against 58\% from scratch and 56\% with kinematics pretraining.
On regrasp, it reaches the first handover in 40\% of trials, against 20\% from scratch and 4\% with kinematics pretraining.
Here kinematics pretraining is \emph{worse} than no pretraining, which suggests that predicting recorded kinematics can teach the trunk habits that hurt downstream performance; we return to this in \cref{sec:analysis}.
Full-task completion remains low for every policy (at most 14\%); even with 2{,}255 demonstrations, regrasp is challenging for all the methods we evaluate.


\subsection{Does pretraining reduce the number of demonstrations needed?}
\label{sec:efficiency}

A main benefit of pretraining is that it reduces how much labeled data a downstream task needs~\cite{henaff2020data,chen2020big}.
We fine-tune each policy on 50\% of the demonstrations for each task.
All policies use the same subset, and its validation demonstrations are unchanged.
We compare training from scratch, kinematics pretraining, and \method, and keep every other setting identical, including the 100k fine-tuning steps.
The three policies are fit with the ACT head, on the same configurations and checkpoints as in \cref{tab:main_results}.

\begin{figure}[t]
  \centering
  \begin{tikzpicture}[font=\scriptsize]
    \begin{groupplot}[
        group style={group size=2 by 1, horizontal sep=1.05cm},
        width=0.37\linewidth, height=2.45cm,
        scale only axis,
        xmin=-0.3, xmax=1.3,
        xtick={0,1}, xticklabels={50\%,100\%},
        axis x line*=bottom, axis y line*=left, y axis line style={draw=none},
        ymajorgrids, grid style={effGrid, line width=0.4pt},
        tick style={draw=none},
        tick label style={font=\scriptsize, text=effMuted},
        title style={font=\footnotesize\bfseries, anchor=south west, at={(0,1)}, xshift=-4pt, yshift=-2pt},
        every axis plot/.append style={line width=1.1pt},
        clip=false,
      ]
      \nextgroupplot[
          title={Pickup success (\%)},
          ymin=0, ymax=100, ytick={0,25,50,75,100},
          yticklabels={0,25,50,75,100},
          legend to name=effLegend,
          legend columns=4,
          legend style={draw=none, font=\scriptsize, inner sep=0pt,
                        /tikz/every odd column/.append style={column sep=1pt},
                        /tikz/every even column/.append style={column sep=5pt}},
          legend image post style={xscale=0.75},
        ]
        \addplot[effMuted, line width=0.6pt, densely dotted, forget plot] coordinates {(-0.3,54) (1.3,54)};
        \addplot[effScratch, dashed, mark=*, mark options={solid, fill=effScratch, draw=white, line width=0.5pt}, mark size=2.1pt]
          coordinates {(0,28) (1,54)};
        \addlegendentry{Scratch}
        \addplot[effKin, mark=square*, mark options={fill=effKin, draw=white, line width=0.5pt}, mark size=1.9pt]
          coordinates {(0,26) (1,60)};
        \addlegendentry{Kinematics}
        \addplot[effFlow, mark=diamond*, mark options={fill=effFlow, draw=white, line width=0.5pt}, mark size=2.8pt]
          coordinates {(0,52) (1,90)};
        \addlegendentry{Flow (ours)}
        \addlegendimage{effMuted, line width=0.6pt, densely dotted}
        \addlegendentry{Scratch, all demos}
        \node[font=\scriptsize\bfseries, above=2pt] at (axis cs:0,52) {52};
        \node[font=\scriptsize\bfseries, above=2pt] at (axis cs:1,90) {90};
      \nextgroupplot[
          title={Regrasp mean stages},
          ymin=0, ymax=1.6, ytick={0,0.5,1,1.5},
          yticklabels={0.0,0.5,1.0,1.5},
        ]
        \addplot[effMuted, line width=0.6pt, densely dotted] coordinates {(-0.3,0.34) (1.3,0.34)};
        \addplot[effScratch, dashed, mark=*, mark options={solid, fill=effScratch, draw=white, line width=0.5pt}, mark size=2.1pt]
          coordinates {(0,0.48) (1,0.34)};
        \addplot[effKin, mark=square*, mark options={fill=effKin, draw=white, line width=0.5pt}, mark size=1.9pt]
          coordinates {(0,0.16) (1,0.60)};
        \addplot[effFlow, mark=diamond*, mark options={fill=effFlow, draw=white, line width=0.5pt}, mark size=2.8pt]
          coordinates {(0,0.98) (1,1.42)};
        \node[font=\scriptsize\bfseries, above=2pt] at (axis cs:0,0.98) {0.98};
        \node[font=\scriptsize\bfseries, above=2pt] at (axis cs:1,1.42) {1.42};
    \end{groupplot}
    \node[anchor=south] at ($(group c1r1.north)!0.5!(group c2r1.north) + (0,0.55cm)$) {\pgfplotslegendfromname{effLegend}};
    \node[anchor=north, text=effMuted] at ($(group c1r1.south)!0.5!(group c2r1.south) + (0,-0.45cm)$) {Fraction of demonstrations};
  \end{tikzpicture}
  \caption{\textbf{Data efficiency.} Real-robot performance with 50\% and 100\% of the demonstrations (ACT head, 50 trials per point). With half the demonstrations, \method matches the pickup success of a policy trained from scratch on all of them and far exceeds it on regrasp, while kinematics pretraining is no better than training from scratch.}
  \label{fig:efficiency}
\end{figure}
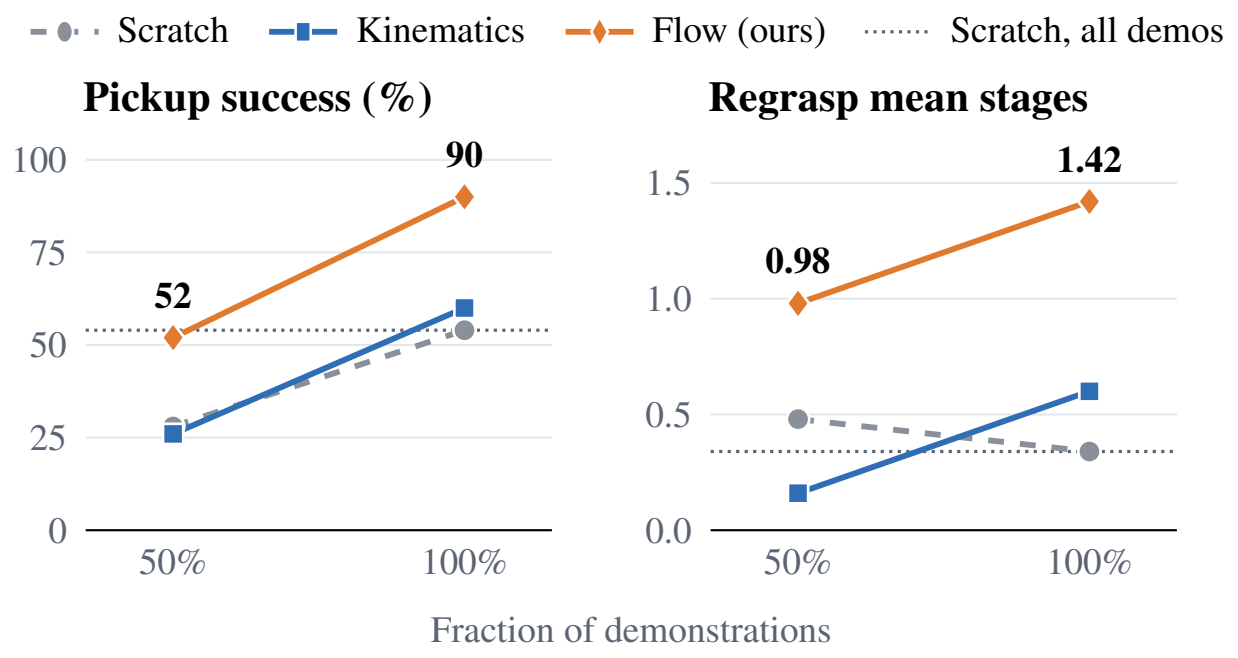

\Cref{fig:efficiency} shows the results.
With half the demonstrations, \method reaches 52\% pickup success, compared with 28\% from scratch and 26\% with kinematics pretraining, and matches a policy trained from scratch on all of the demonstrations (54\%).
The gap is larger on regrasp, the longer task.
With 50\% of the demonstrations, \method reaches the first handover in 32\% of trials, compared with 4\% for training from scratch on all of them (0.98 vs.\ 0.34 mean stages).
Kinematics pretraining, in contrast, loses most of its benefit with fewer demonstrations (0.60 to 0.16 mean stages on regrasp) and is no better than training from scratch on either task.
Scene-flow pretraining can thus replace half of the expert demonstrations without loss on pickup and with a large gain on regrasp.

\subsection{What in the video helps?}
\label{sec:ablation}

The previous sections show that scene-flow pretraining helps.
We now ask which part of the target matters.
\Cref{tab:ablation} compares eight pretraining targets, all computed from the same video anchors and tracked points, so that only the target changes.
Each pretrained trunk is fine-tuned on needle pickup with the ACT head.
To evaluate this many models on the robot, we roll out two checkpoints (40k and 60k steps) per model, for 20 trials each; \cref{tab:main_results} indicates that pickup success is a good proxy, since the policies that do well on pickup also do well on regrasp.

The targets vary along three axes, each changing one factor at a time.
The first is whether the model also predicts kinematics: kinematics only, kinematics plus metric 3D flow, and metric 3D flow only (\method).
The second is how the motion is represented, with kinematics held fixed: alongside kinematics, we predict metric 3D flow, 2D pixel tracks (no depth), 3D flow divided by the median scene depth (scale-free, as from a monocular depth estimate), or depth only (3D positions, no motion).
The third is the type of pretraining flow head: L1 regression or flow matching, applied to both kinematics plus flow and flow only.

\para{Metric 3D motion is the most useful target we tested.}
With kinematics held fixed, metric 3D flow gives the highest success (80\%), ahead of 2D pixel tracks (70\%) and depth alone (60\%), neither of which improves on no pretraining; we note that at $n=20$, these gaps are suggestive rather than conclusive.
Removing metric scale is clearly harmful: with scale-free flow, success drops to 35\%.
On held-out video, the scale-free flow head barely improved on predicting no motion, so without metric scale the target carries little usable signal.
We did not test these alternatives without kinematics, so we cannot rule out that they would help on their own.

\para{Kinematics do not help, and may hurt.}
Predicting kinematics alone gives the same success as no pretraining (70\%).
Adding kinematics to metric 3D flow lowers success from 100\% to 80\% compared with flow alone.
The best target uses no robot data at all, which is also what makes it applicable to calibrated clinical stereo video; we use logged kinematics only to select static-camera clips, which a visual camera-motion check could replace.

\para{A generative flow head does not help during pretraining.}
Replacing the L1 head with a flow-matching head to predict 3D scene flow reduces success in both settings, and first-attempt success in particular (55\% to 10\% for flow only).
For pretraining the trunk, a simple regression target is preferable.
This concerns only the \emph{pretraining} flow head; the flow-matching \emph{action} head used for fine-tuning in \cref{sec:main} works well on top of a trunk pretrained with L1.

\begin{table}[t]
  \centering\small
  \setlength{\tabcolsep}{3.5pt}
  \begin{tabular}{@{}lcc@{}}
    \toprule
    Pretraining target & Success & 1st attempt \\
    \midrule
    None (from scratch)                    & 70\% & 45\% \\
    Kinematics                             & 70\% & 50\% \\
    Kinematics + 2D tracks                 & 70\% & 45\% \\
    Kinematics + depth only                & 60\% & 30\% \\
    Kinematics + scale-free 3D flow        & 35\% & 20\% \\
    Kinematics + metric 3D flow            & 80\% & 45\% \\
    \quad with flow-matching flow head     & 65\% & 20\% \\
    \textbf{Metric 3D flow only (\method)} & \textbf{100\%} & \textbf{55\%} \\
    \quad with flow-matching flow head     & 55\% & 10\% \\
    \bottomrule
  \end{tabular}
  \caption{\textbf{Pretraining targets.} Needle pickup success within three attempts and on the first attempt, after fine-tuning each pretrained trunk with the ACT head. $n = 20$ trials per target (10 needle configurations $\times$ the 40k and 60k checkpoints). All targets are computed from the same anchors and points. ``Flow-matching flow head'': the pretraining flow head is generative instead of L1 (\cref{sec:model}); the fine-tuned policy is unchanged. Different session from \cref{tab:main_results}.}
  \label{tab:ablation}
\end{table}
\section{Why does scene flow pretraining help?}
\label{sec:analysis}

\Cref{sec:exp} shows that scene-flow pretraining helps and that kinematics pretraining does not. In this section we ask \emph{why}, by looking inside the policies.

Prior work links rollout success in manipulation to visual understanding: how well a model captures the task-relevant objects in a scene~\cite{burns2023makes,wen2025datacentric}. In \method, however, the visual tokenizer (DINOv2) is shared by every policy and frozen, so it cannot explain the differences between them. We therefore analyze what the \emph{policy trunk} encodes: the output of the ACT encoder at the image tokens, which is what the action queries (and, during pretraining, the point queries) read from. We compare the trunks right after pretraining and after fine-tuning, on frames from held-out demonstrations.

\para{Dense linear probes.}
To measure how much scene geometry a trunk encodes, we train a linear probe on its frozen features to predict needle masks, instrument masks, and stereo disparity (details in~\cref{sec:supp_probes}).
We report the relative depth error (abs\_rel) on needle pixels, instrument pixels, and the whole image, and the segmentation IoU for needle and instruments, on 80 held-out demonstrations (4.5k frames).
Depth labels come from Fast-FoundationStereo~\cite{wen2026fast}, the same model used to build the scene-flow pretraining targets, so flow pretraining could in principle favor these labels.
However, the probe frames are robot demonstrations, for which the trunks never saw stereo labels during pretraining.
For reference, we also probe the frozen DINOv2 tokens that every trunk takes as input; these are wider (1024-d instead of 512-d), limiting direct comparison.
\begin{table}[t]
  \centering\footnotesize
  \setlength{\tabcolsep}{3pt}
  \begin{tabular}{@{}lccccc@{}}
    \toprule
    & \multicolumn{3}{c}{Depth abs\_rel $\downarrow$} & \multicolumn{2}{c}{IoU $\uparrow$} \\
    \cmidrule(lr){2-4}\cmidrule(l){5-6}
    Features & needle & instr. & image & needle & instr. \\
    \midrule
    DINOv2-IH (1024-d)       & 0.10 & 0.08 & 0.04 & 0.49 & 0.95 \\
    \midrule
    \multicolumn{6}{@{}l}{\emph{Pretrained trunk (512-d)}} \\
    Kinematics            & 0.13 & 0.15 & 0.06 & 0.44 & 0.93 \\
    Kin.\ + flow          & 0.10 & 0.11 & 0.04 & 0.44 & 0.94 \\
    Flow only (ours)      & \textbf{0.08} & \textbf{0.09} & \textbf{0.04} & \textbf{0.46} & \textbf{0.95} \\
    \midrule
    \multicolumn{6}{@{}l}{\emph{Fine-tuned policy (512-d)}} \\
    None (from scratch)   & 0.14 & 0.14 & 0.07 & 0.43 & 0.93 \\
    Kinematics            & 0.15 & 0.15 & 0.06 & 0.43 & 0.92 \\
    Kin.\ + flow          & 0.13 & \textbf{0.11} & \textbf{0.05} & 0.43 & 0.93 \\
    Flow only (ours)      & \textbf{0.11} & 0.12 & 0.06 & \textbf{0.44} & \textbf{0.94} \\
    \bottomrule
  \end{tabular}
  \caption{\textbf{What the policy trunk encodes.} Linear probes on 80 held-out demonstrations. Bold: best trunk in each block. Over 2{,}000 bootstrap resamples of the validation demonstrations, flow-only pretraining gives the lowest needle and instrument depth error among pretrained trunks.}
  \label{tab:probes}
\end{table}

\para{Flow pretraining builds metric needle geometry, finetuning maintains it.}
Among the pretrained trunks, the flow-only trunk of \method encodes the scene best on every probe (\cref{tab:probes}).
It is also the only trunk that encodes the needle's depth \emph{better} than the DINOv2 features it takes as input (0.08 vs.\ 0.10 abs\_rel), even though it has half as many channels.
Pretraining with kinematics has the opposite effect: kinematics-only pretraining encodes needle depth worse than its own input (0.13 vs.\ 0.10), while adding flow restores it to the input's level.

Fine-tuning on the pickup demonstrations removes some of this information from every trunk, but the ranking remains consistent.
After fine-tuning, the \method policy still encodes the needle's depth best (0.11), ahead of kin.\ + flow (0.13) and of scratch and kinematics, which are tied (0.14 and 0.15).
The ranking on needle depth (flow only $>$ kin.\ + flow $>$ scratch and kinematics) is also the ranking of real-robot success in \cref{tab:main_results,tab:ablation}, whereas offline action error on held-out demonstrations is nearly identical for all four policies (\cref{sec:supp_offline}).
Consistent with prior work~\cite{burns2023makes}, how well a representation captures the task-relevant object is associated with how well the policy performs across the four initializations we compare.

\para{Kinematics pretraining teaches the trunk to rely on the robot state rather than the scene.}
Besides the image tokens, ACT's encoder receives a token for the robot state, the current end-effector pose of both instruments.
Imitation policies that predict actions from such proprioceptive input are known to rely on it. Although the detriment is indistinguishable offline, these models fail at rollout and closed loop~\cite{de2019causal,wen2020fighting}.
We therefore measure how much of the image tokens' attention goes to the state token (\cref{sec:supp_state}).
After kinematics pretraining, the image tokens give about a quarter of their attention to it (24.5\%), and the habit survives fine-tuning (21.3\%, compared with about 8\% for scratch and \method).
Flow pretraining cannot learn this shortcut, since its state input is zero. We therefore attribute the weakness of kinematics pretraining to this shortcut rather than to the kinematic target itself; pretraining on kinematics with the state masked is a control we leave to future work.

\section{Conclusion}
\label{sec:conclusion}

We present \method, a method that pretrains surgical policy networks on clinical video by forecasting how points in the scene move in metric 3D.
Fine-tuned and evaluated on a minimally invasive surgical robot, \method improves needle pickup and needle regrasp over training from scratch, with both a deterministic and a generative action head.
The pretraining target matters: metric 3D motion helps, while the recorded kinematics do not.
Flow pretraining builds a trunk that encodes the needle's 3D geometry, whereas kinematics pretraining teaches it to read the robot state instead of the scene.

\para{Limitations.}
We evaluate on a single platform and on suture pads; transfer to other robots and to ex vivo or in vivo tissue remain as future work.
Our labels are pseudo-labels that inherit the errors of the stereo and optical-flow models, and absolute success on the full regrasp task is still low.
We compare pretraining targets within one architecture, not against other video-pretraining systems such as SurgVIL or Surgical WAM, whose data and platforms differ from ours.
We see \method as a first step toward policies that learn from the vast record of surgeries already performed.

\para{LLM Usage Statement.} The authors used Claude Opus 5 to assist with the writing of this manuscript; authors wrote the initial draft and iteratively improved with the help of AI throughout the manuscript. The authors also employed the same AI model for code writing and figure generation.

{
    \small
    \bibliographystyle{ieeenat_fullname}
    \bibliography{main}

@inproceedings{zhao2023act,
  title={Learning Fine-Grained Bimanual Manipulation with Low-Cost Hardware},
  author={Zhao, Tony Z. and Kumar, Vikash and Levine, Sergey and Finn, Chelsea},
  booktitle={Robotics: Science and Systems (RSS)}, year={2023}}

@inproceedings{chi2023diffusion,
  title={Diffusion Policy: Visuomotor Policy Learning via Action Diffusion},
  author={Chi, Cheng and Feng, Siyuan and Du, Yilun and Xu, Zhenjia and Cousineau, Eric and Burchfiel, Benjamin and Song, Shuran},
  booktitle={Robotics: Science and Systems (RSS)}, year={2023}}

@article{black2024pi0,
  title={$\pi_0$: A Vision-Language-Action Flow Model for General Robot Control},
  author={Black, Kevin and others}, journal={arXiv preprint arXiv:2410.24164}, year={2024}}

@article{intelligence2025pi05,
  title={$\pi_{0.5}$: a Vision-Language-Action Model with Open-World Generalization},
  author={{Physical Intelligence} and others}, journal={arXiv preprint arXiv:2504.16054}, year={2025}}

@article{oquab2024dinov2,
  title={{DINOv2}: Learning Robust Visual Features without Supervision},
  author={Oquab, Maxime and others}, journal={Transactions on Machine Learning Research}, year={2024}}

@inproceedings{o2024open,
  title={Open x-embodiment: Robotic learning datasets and rt-x models: Open x-embodiment collaboration 0},
  author={O’Neill, Abby and Rehman, Abdul and Maddukuri, Abhiram and Gupta, Abhishek and Padalkar, Abhishek and Lee, Abraham and Pooley, Acorn and Gupta, Agrim and Mandlekar, Ajay and Jain, Ajinkya and others},
  booktitle={2024 IEEE International Conference on Robotics and Automation (ICRA)},
  pages={6892--6903},
  year={2024},
  organization={IEEE}
}

@article{han2026scared,
  title={SCARED-C: Corrected Camera Poses for Endoscopic Depth Estimation},
  author={Han, John J and Schmidt, Adam and Allan, Max and Wu, Jie Ying and Mohareri, Omid},
  journal={arXiv preprint arXiv:2605.16628},
  year={2026}
}

@inproceedings{liang2024real,
  title={Real-to-sim deformable object manipulation: Optimizing physics models with residual mappings for robotic surgery},
  author={Liang, Xiao and Liu, Fei and Zhang, Yutong and Li, Yuelei and Lin, Shan and Yip, Michael},
  booktitle={2024 IEEE international conference on robotics and automation (ICRA)},
  pages={15471--15477},
  year={2024},
  organization={IEEE}
}

@inproceedings{ross2011reduction,
  title={A reduction of imitation learning and structured prediction to no-regret online learning},
  author={Ross, St{\'e}phane and Gordon, Geoffrey and Bagnell, Drew},
  booktitle={Proceedings of the fourteenth international conference on artificial intelligence and statistics},
  pages={627--635},
  year={2011},
  organization={JMLR Workshop and Conference Proceedings}
}

@article{he2025cosmos,
  title={Cosmos-H-Surgical: Learning Surgical Robot Policies from Videos via World Modeling},
  author={He, Yufan and Guo, Pengfei and Xu, Mengya and Li, Zhaoshuo and Myronenko, Andriy and Imans, Dillan and Liu, Bingjie and Yang, Dongren and Gu, Mingxue and Ji, Yongnan and others},
  journal={arXiv preprint arXiv:2512.23162},
  year={2025}
}

@article{bao2026surgical,
  title={Surgical WAM: A World-Action Model for Data-Efficient Surgical Robot Learning},
  author={Bao, Wenrui and Jiang, Tianyun and Chen, Zhiben and Lim, Ser-Nam and Peng, Peter D and Shang, Yuzhang},
  journal={arXiv preprint arXiv:2608.11204},
  year={2026}
}

@article{chen2026surgvil,
  title={SurgVIL: Scaling Surgical Robot Imitation Learning with Open-source Surgical Videos},
  author={Chen, Xinhao and Chen, JuoTung and Nelson, Nigel and Goldenberg, Antony and Haworth, Jesse and Huver, Sean D and Krieger, Axel},
  journal={arXiv preprint arXiv:2608.16058},
  year={2026}
}

@article{scheikl2022sim,
  title={Sim-to-real transfer for visual reinforcement learning of deformable object manipulation for robot-assisted surgery},
  author={Scheikl, Paul Maria and Tagliabue, Eleonora and Gyenes, Bal{\'a}zs and Wagner, Martin and Dall'Alba, Diego and Fiorini, Paolo and Mathis-Ullrich, Franziska},
  journal={IEEE Robotics and Automation Letters},
  volume={8},
  number={2},
  pages={560--567},
  year={2022},
  publisher={IEEE}
}

@inproceedings{ye2016real,
  title={Real-time 3d tracking of articulated tools for robotic surgery},
  author={Ye, Menglong and Zhang, Lin and Giannarou, Stamatia and Yang, Guang-Zhong},
  booktitle={International conference on medical image computing and computer-assisted intervention},
  pages={386--394},
  year={2016},
  organization={Springer}
}

@inproceedings{han2026endopbr,
  title={EndoPBR: Photorealistic Synthetic Data for Surgical 3D Vision via Physically-based Rendering},
  author={Han, John J and Wu, Jie Ying},
  booktitle={2026 IEEE/CVF Winter Conference on Applications of Computer Vision (WACV)},
  pages={5601--5611},
  year={2026},
  organization={IEEE}
}

@article{de2025scaling,
  title={Scaling up self-supervised learning for improved surgical foundation models},
  author={de Jong, Ronald and Carolus, HJ and Franciscus, HA and van Jaarsveld, Romy C and van Hillegersberg, Richard and Josien, PW and de With, Peter HN and al Khalil, Yasmina and van Der Sommen, Fons and others},
  journal={Medical Image Analysis},
  pages={103873},
  year={2025},
  publisher={Elsevier}
}

@inproceedings{wen2026fast,
  title={Fast-FoundationStereo: Real-time zero-shot stereo matching},
  author={Wen, Bowen and Dewan, Shaurya and Birchfield, Stan},
  booktitle={Proceedings of the IEEE/CVF conference on computer vision and pattern recognition},
  pages={7513--7524},
  year={2026}
}

@article{perez2026surglavi,
  title={Surglavi: Large-scale hierarchical dataset for surgical vision--language representation learning},
  author={Perez, Alejandra and Nwoye, Chinedu and Kermani, Ramtin Raji and Mohareri, Omid and Jamal, Muhammad Abdullah},
  journal={Medical Image Analysis},
  pages={103982},
  year={2026},
  publisher={Elsevier}
}

@misc{abc2026,
  title         = {Scalable Behavior Cloning with Open Data, Training, and Evaluation},
  author        = {Arthur Allshire and Himanshu Gaurav Singh and Ritvik Singh and Adam Rashid and Hongsuk Choi and David McAllister and Justin Yu and Yiyuan Chen and Huang Huang and Pieter Abbeel and Xi Chen and Rocky Duan and Phillip Isola and Jitendra Malik and Fred Shentu and Guanya Shi and Philipp Wu and Angjoo Kanazawa},
  year          = {2026},
  eprint        = {2606.27375},
  archivePrefix = {arXiv},
  primaryClass  = {cs.RO},
  doi           = {10.48550/arXiv.2606.27375},
  url           = {https://arxiv.org/abs/2606.27375},
}

@article{chen2020big,
  title={Big self-supervised models are strong semi-supervised learners},
  author={Chen, Ting and Kornblith, Simon and Swersky, Kevin and Norouzi, Mohammad and Hinton, Geoffrey E},
  journal={Advances in neural information processing systems},
  volume={33},
  pages={22243--22255},
  year={2020}
}

@inproceedings{henaff2020data,
  title={Data-efficient image recognition with contrastive predictive coding},
  author={Henaff, Olivier},
  booktitle={International conference on machine learning},
  pages={4182--4192},
  year={2020},
  organization={PMLR}
}

@article{hu2021lora,
  title={Lora: Low-rank adaptation of large language models},
  author={Hu, Edward J and Shen, Yelong and Wallis, Phillip and Allen-Zhu, Zeyuan and Li, Yuanzhi and Wang, Shean and Wang, Lu and Chen, Weizhu},
  journal={arXiv preprint arXiv:2106.09685},
  year={2021}
}

@inproceedings{carion2020end,
  title={End-to-end object detection with transformers},
  author={Carion, Nicolas and Massa, Francisco and Synnaeve, Gabriel and Usunier, Nicolas and Kirillov, Alexander and Zagoruyko, Sergey},
  booktitle={European conference on computer vision},
  pages={213--229},
  year={2020},
  organization={Springer}
}

@article{han2026dart,
  title={DART: Depth-as-Target Pretraining for Surgical Vision Foundation Models},
  author={Han, John J and Schmidt, Adam and Jamal, Muhammad Abdullah and Wu, Jie Ying and Mohareri, Omid},
  journal={arXiv preprint arXiv:2609.04555},
  year={2026}
}

@inproceedings{devlin2019bert,
  title={Bert: Pre-training of deep bidirectional transformers for language understanding},
  author={Devlin, Jacob and Chang, Ming-Wei and Lee, Kenton and Toutanova, Kristina},
  booktitle={Proceedings of the 2019 conference of the North American chapter of the association for computational linguistics: human language technologies, volume 1 (long and short papers)},
  pages={4171--4186},
  year={2019}
}

@article{radford2018improving,
  title={Improving language understanding by generative pre-training},
  author={Radford, Alec and Narasimhan, Karthik and Salimans, Tim and Sutskever, Ilya and others},
  year={2018},
  publisher={San Francisco, CA, USA}
}

@article{radford2019language,
  title={Language models are unsupervised multitask learners},
  author={Radford, Alec and Wu, Jeffrey and Child, Rewon and Luan, David and Amodei, Dario and Sutskever, Ilya and others},
  journal={OpenAI blog},
  volume={1},
  number={8},
  pages={9},
  year={2019}
}

@article{brohan2023rt,
  title={Rt-2: Vision-language-action models transfer web knowledge to robotic control},
  author={Brohan, Anthony and Brown, Noah and Carbajal, Justice and Chebotar, Yevgen and Chen, Xi and Choromanski, Krzysztof and Ding, Tianli and Driess, Danny and Dubey, Avinava and Finn, Chelsea and others},
  journal={arXiv preprint arXiv:2307.15818},
  year={2023}
}

@article{brohan2022rt,
  title={Rt-1: Robotics transformer for real-world control at scale},
  author={Brohan, Anthony and Brown, Noah and Carbajal, Justice and Chebotar, Yevgen and Dabis, Joseph and Finn, Chelsea and Gopalakrishnan, Keerthana and Hausman, Karol and Herzog, Alex and Hsu, Jasmine and others},
  journal={arXiv preprint arXiv:2212.06817},
  year={2022}
}

@inproceedings{he2022masked,
  title={Masked autoencoders are scalable vision learners},
  author={He, Kaiming and Chen, Xinlei and Xie, Saining and Li, Yanghao and Doll{\'a}r, Piotr and Girshick, Ross},
  booktitle={2022 IEEE/CVF conference on computer vision and pattern recognition (CVPR)},
  pages={15979--15988},
  year={2022},
  organization={IEEE}
}

@inproceedings{wen2024atm,
  title={Any-point Trajectory Modeling for Policy Learning},
  author={Wen, Chuan and Lin, Xingyu and So, John and Chen, Kai and Dou, Qi and Gao, Yang and Abbeel, Pieter},
  booktitle={Robotics: Science and Systems (RSS)}, year={2024}}

@inproceedings{wang2024seaflow,
  title={{SEA-RAFT}: Simple, Efficient, Accurate {RAFT} for Optical Flow},
  author={Wang, Yihan and Lipson, Lahav and Deng, Jia}, booktitle={ECCV}, year={2024}}

@inproceedings{kim2024srt,
  title={Surgical Robot Transformer ({SRT}): Imitation Learning for Surgical Tasks},
  author={Kim, Ji Woong and others}, booktitle={Conference on Robot Learning (CoRL)}, year={2024}}

@article{schmidt2025point,
  title={Point Tracking in Surgery--The 2024 Surgical Tattoos in Infrared (STIR) Challenge},
  author={Schmidt, Adam and Karaoglu, Mert Asim and Sinha, Soham and Jang, Mingang and Ha, Ho-Gun and Jung, Kyungmin and Gu, Kyeongmo and Ullah, Ihsan and Lee, Hyunki and {\v{S}}er{\`y}ch, Jon{\'a}{\v{s}} and others},
  journal={arXiv preprint arXiv:2503.24306},
  year={2025}
}

@article{han2026stereopolicy,
  title={Stereopolicy: Improving robotic manipulation policies via stereo perception},
  author={Han, Evans and Jiang, Yunfan and Wang, Yingke and Xiao, Haoyue and Huang, Huang and Xie, Jianwen and Wu, Jiajun and Fei-Fei, Li and Zhang, Ruohan},
  journal={arXiv preprint arXiv:2605.09989},
  year={2026}
}

@article{wen2020fighting,
  title={Fighting copycat agents in behavioral cloning from observation histories},
  author={Wen, Chuan and Lin, Jierui and Darrell, Trevor and Jayaraman, Dinesh and Gao, Yang},
  journal={Advances in Neural Information Processing Systems},
  volume={33},
  pages={2564--2575},
  year={2020}
}

@article{de2019causal,
  title={Causal confusion in imitation learning},
  author={De Haan, Pim and Jayaraman, Dinesh and Levine, Sergey},
  journal={Advances in neural information processing systems},
  volume={32},
  year={2019}
}

@inproceedings{kazanzides2014dvrk,
  title={An Open-Source Research Kit for the da Vinci Surgical System},
  author={Kazanzides, Peter and others}, booktitle={ICRA}, year={2014}}

@inproceedings{lipman2023flow,
  title={Flow Matching for Generative Modeling},
  author={Lipman, Yaron and Chen, Ricky T. Q. and Ben-Hamu, Heli and Nickel, Maximilian and Le, Matt},
  booktitle={ICLR}, year={2023}}

@article{hung20263pointr,
  title={3PoinTr: From Human Videos to Robot Policies with 3D Point-Track Plans}, 
  author={Hung, Adam and Duisterhof, Bardienus Pieter and Ichnowski, Jeffrey},
  journal={arXiv preprint arXiv:2603.08485},
  year={2026}
}

@inproceedings{peebles2023dit,
  title={Scalable Diffusion Models with Transformers},
  author={Peebles, William and Xie, Saining}, booktitle={ICCV}, year={2023}}

@article{niu2025arm4r,
  title={Pre-training auto-regressive robotic models with 4d representations},
  author={Niu, Dantong and Sharma, Yuvan and Xue, Haoru and Biamby, Giscard and Zhang, Junyi and Ji, Ziteng and Darrell, Trevor and Herzig, Roei},
  journal={arXiv preprint arXiv:2502.13142},
  year={2025}
}

@inproceedings{zhu2025spa,
  title={Spa: 3d spatial-awareness enables effective embodied representation},
  author={Zhu, Haoyi and Yang, Honghui and Wang, Yating and Yang, Jiange and Wang, Limin and He, Tong},
  booktitle={International Conference on Learning Representations},
  volume={2025},
  pages={26361--26391},
  year={2025}
}

@inproceedings{lee2026tracegen,
  title={Tracegen: World modeling in 3d trace space enables learning from cross-embodiment videos},
  author={Lee, Seungjae and Jung, Yoonkyo and Chun, Inkook and Lee, Yao-Chih and Cai, Zikui and Huang, Hongjia and Talreja, Aayush and Dao, Tan and Liang, Yongyuan and Huang, Jia-Bin and others},
  booktitle={Proceedings of the IEEE/CVF Conference on Computer Vision and Pattern Recognition},
  pages={20721--20731},
  year={2026}
}

@article{duisterhof2026pointzero,
  title={PointZero: 3D Point Track Completion for Learning Transferable 3D Dynamics},
  author={Duisterhof, Bardienus P and Zhang, Kaifeng and Hung, Adam and Wen, Bowen and Birchfield, Stan and Li, Yunzhu and Ramanan, Deva and Ichnowski, Jeffrey},
  journal={arXiv preprint arXiv:2609.19142},
  year={2026}
}

@article{chen2026surgflow,
  title={{SurgFlow}: {3D} Object-Centric Contact Flow for Surgical Robot Manipulation},
  author={Chen, Changwei and Liang, Xiao and Yang, Yinuo and Shen, Nicole and Zhang, Peihan and Wickenhiser, Sara and Liang, Zekai and Atar, Soofiyan and Yip, Michael},
  journal={arXiv preprint arXiv:2609.33237}, year={2026}}

@article{simeoni2025dinov3,
  title={Dinov3},
  author={Sim{\'e}oni, Oriane and Vo, Huy V and Seitzer, Maximilian and Baldassarre, Federico and Oquab, Maxime and Jose, Cijo and Khalidov, Vasil and Szafraniec, Marc and Yi, Seungeun and Ramamonjisoa, Micha{\"e}l and others},
  journal={arXiv preprint arXiv:2508.10104},
  year={2025}
}

@article{burns2023makes,
  title={What Makes Pre-Trained Visual Representations Successful for Robust Manipulation?},
  author={Burns, Kaylee and Witzel, Zach and Hamid, Jubayer Ibn and Yu, Tianhe and Finn, Chelsea and Hausman, Karol},
  journal={arXiv preprint arXiv:2312.12444}, year={2023}}

@inproceedings{wen2025datacentric,
  title={A Data-Centric Revisit of Pre-Trained Vision Models for Robot Learning},
  author={Wen, Xin and Zhao, Bingchen and Chen, Yilun and Pang, Jiangmiao and Qi, Xiaojuan},
  booktitle={CVPR}, year={2025}}
}

\providecommand{\method}{FLOW-ACT}
\providecommand{\note}[1]{\textcolor{blue}{[#1]}}
\providecommand{\pending}{\textcolor{orange}{pending}}
\providecommand{\figplaceholder}[2]{\fbox{\parbox[c][#1][c]{0.95\linewidth}{\centering\footnotesize #2}}}

\clearpage
\appendix
\renewcommand{\thesection}{\Alph{section}}
\renewcommand{\thetable}{S\arabic{table}}
\renewcommand{\thefigure}{S\arabic{figure}}
\setcounter{table}{0}
\setcounter{figure}{0}

\begin{center}
  {\Large\bfseries Supplementary Material}
\end{center}

\noindent
\Cref{sec:supp_data} describes the clinical video and how we turn it into scene-flow labels.
\Cref{sec:supp_robot} describes the robot, the tasks, and how we evaluate.
\Cref{sec:supp_encoder,sec:supp_training,sec:supp_heads} give the training details of the DINOv2, of pretraining and fine-tuning, and of the action heads.
\Cref{sec:supp_rollouts} reports the full rollout results, \cref{sec:supp_analysis} additional analysis, and \cref{sec:supp_exps} additional experiments. Citation numbers refer to the reference list of the main paper. 

\section{Clinical video and scene-flow labels}
\label{sec:supp_data}

\subsection{Clinical video pool}
\label{sec:supp_pool}
In collaboration and partnerships with various hospitals, we curate a dataset of recordings from a minimally invasive surgical robotic system containing real surgeries on patients. We sampled 1{,}296 procedure videos; 121 of them yield no anchor that passes the filters (for example, the camera never stays still while an instrument is driven), leaving 1{,}175 procedures, split by video into 1{,}149 training and 26 validation videos so that no validation frame comes from a training procedure.
Of the 2.69M candidate anchors (one every 10 frames with a full 1.7\,s window after it), 1.35M (50\%) pass the anchor filters; the median video keeps 59\% of its candidates.

\paragraph{Ethics and broader impact statement.}
\label{sec:supp_ethics}
Our work concerns pretraining methods for surgical robot autonomy, a safety-critical domain in which policy errors could harm patients.
All robot experiments in this paper are performed on suture pads in a laboratory setting; no policy was deployed on humans or animals.
\method is not a clinical system: the success rates we report, while substantially improved, remain far below what clinical use would require, and any translation would require extensive validation, and regulatory approval.

The clinical videos were recorded during routine robot-assisted procedures, with patient consent for the use of their data, and were obtained under consent agreements overseen by an institutional review board or equivalent body; the data cannot be released.
All recordings are de-identified.
The institution and review body will be named upon acceptance, to preserve anonymity.

\subsection{Label generation}
\label{sec:supp_labels}
Each training clip spans 100 frames: an anchor frame and ten keyframes spaced 10 frames apart after it.
Depth comes from Fast-FoundationStereo~\cite{wen2026fast} on the rectified stereo pair, and 2D motion from SEA-RAFT~\cite{wang2024seaflow} between consecutive keyframes. Since the recordings are natively rectified, each frame's stereo calibration converts disparity to metric depth.
Points are chained through the optical flow from the anchor to each keyframe and lifted to 3D with the depth at each keyframe, giving each point's metric 3D position at the anchor, $P_0$, and its displacement at each keyframe. We use H200 GPUs for pseudolabel generation and save labels on disk at $518\times 518$ in \verb|hdf5| format. 

\subsection{Filtering}
\label{sec:supp_filters}
We keep a clip only if the camera is static and at least one instrument is driven, and a point's track only while its depth and flow pass the consistency checks in \cref{tab:supp_filters}.
A track that fails at one keyframe is dropped from that keyframe onwards.

\begin{table}[h]
  \centering\footnotesize
  \setlength{\tabcolsep}{3pt}
  \begin{tabular}{@{}llp{2.9cm}@{}}
    \toprule
    Stage & Test & Threshold \\
    \midrule
    Clip   & Camera static                      & $<0.5$\,mm and $<0.5^\circ$ over the clip \\
           & An instrument is driven            & $\ge 1$ arm for the whole window \\
           & Kinematics readable                & all frames \\
    \midrule
    Depth  & Left--right consistency            & $<1$\,px \\
           & Brightness                         & $>0.03$ \\
           & Flying pixels                      & 3\% \\
           & Depth range                        & 5--300\,mm \\
    \midrule
    Flow   & Fwd--bwd consistency              & $<1$\,px $+\,5\%\,|f|$ (half res.) \\
           & Photometric change                 & $<0.12$ \\
           & Depth edge                         & 5\% \\
           & Depth change across keyframes      & $|\Delta Z| < 25\%$ \\
    \midrule
    Anchor & Surviving points                   & $\ge 16$ \\
    \bottomrule
  \end{tabular}
  \caption{\textbf{Label filters.} We apply numerous filtering steps to ensure training data quality. We only use clips where the camera is static, at least one instrument is being driven, and kinematic signals are valid. We convert disparity maps in both $L \leftrightarrow$ R directions and only use pixels where they agree up to a pixel. We also ignore dark pixels and only use depth values between 5-300\,mm. We apply similar checks for optical flow.}
  \label{tab:supp_filters}
\end{table}

\subsection{Point sampling}
\label{sec:supp_points}
We store 512 points per anchor: 256 sampled uniformly over the valid pixels and 256 sampled in proportion to their motion.
During pretraining, each anchor contributes 256 point queries, up to half of them drawn from points that move more than 5\,mm and the rest uniformly.
 
\begin{figure}[t]
  \centering
  \includegraphics[width=\linewidth]{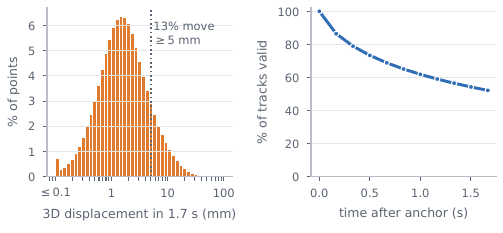}
  \caption{\textbf{Label statistics.} Left: 3D displacement of uniformly sampled points over the 1.7\,s horizon (dotted: the 5\,mm mover threshold). Right: fraction of tracked points still valid at each keyframe.}
  \label{fig:supp_label_stats}
\end{figure}

\paragraph{How much of the scene moves.}
\Cref{fig:supp_label_stats} (left) shows the 3D displacement of the uniformly sampled points over the 1.7\,s horizon, measured on 232k anchors (200 per video).
The median point moves 1.6\,mm, and only 12.8\% move 5\,mm or more (3.8\% move 10\,mm or more).
Motion is nevertheless present almost everywhere: 92\% of anchors contain at least one such mover, and in 46\% of anchors at least a tenth of the points move.
This imbalance is why we sample points in proportion to their motion: among the 512 stored points, the median anchor has 108 movers, and 42\% of anchors have enough to fill all 128 mover queries.
 
\paragraph{Track survival.}
A track ends at the first keyframe where it fails a check.
86\% of tracked points are still valid after one keyframe (0.17\,s) and 52\% at the end of the horizon (\cref{fig:supp_label_stats}, right).
 
\begin{figure*}[t]
  \centering
  \includegraphics[width=\linewidth]{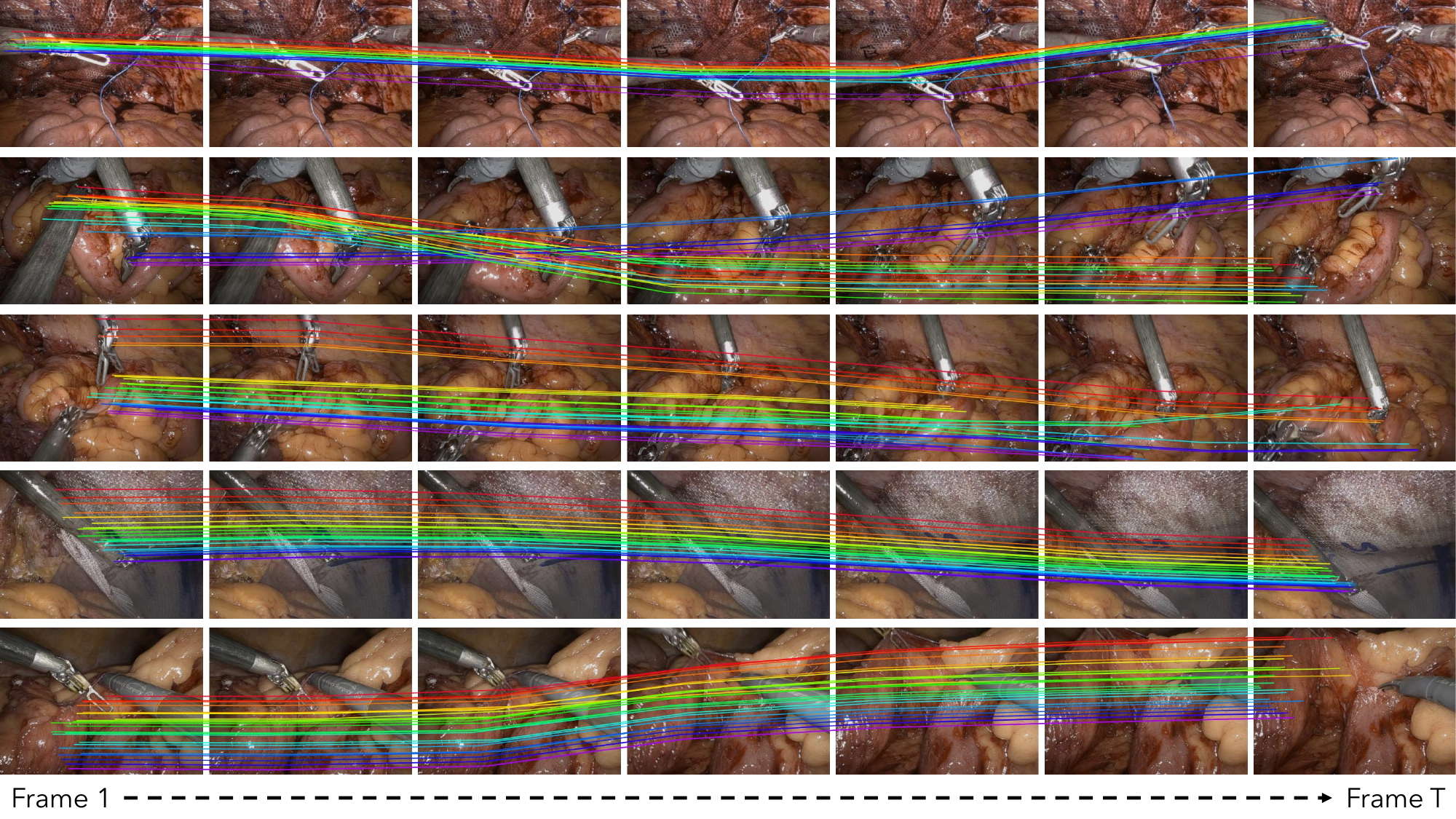}
  \caption{\textbf{Samples of scene-flow pseudolabels.} Tracks of the moving points over frames. For the purposes of visualization, we only draw the top 25 pixels that move more than $20$\,mm. In reality, our labeling pipeline stores 512 tracks per clip; during training, \method samples 256. }
  \label{fig:supp_flow_examples}
\end{figure*}
 
\section{Robot setup, tasks, and evaluation}
\label{sec:supp_robot}
 
\subsection{Robot setup}
\label{sec:supp_setup}
\begin{figure}[t]
    \centering
    \begin{subfigure}[b]{0.48\linewidth}
        \centering
        \includegraphics[width=\linewidth]{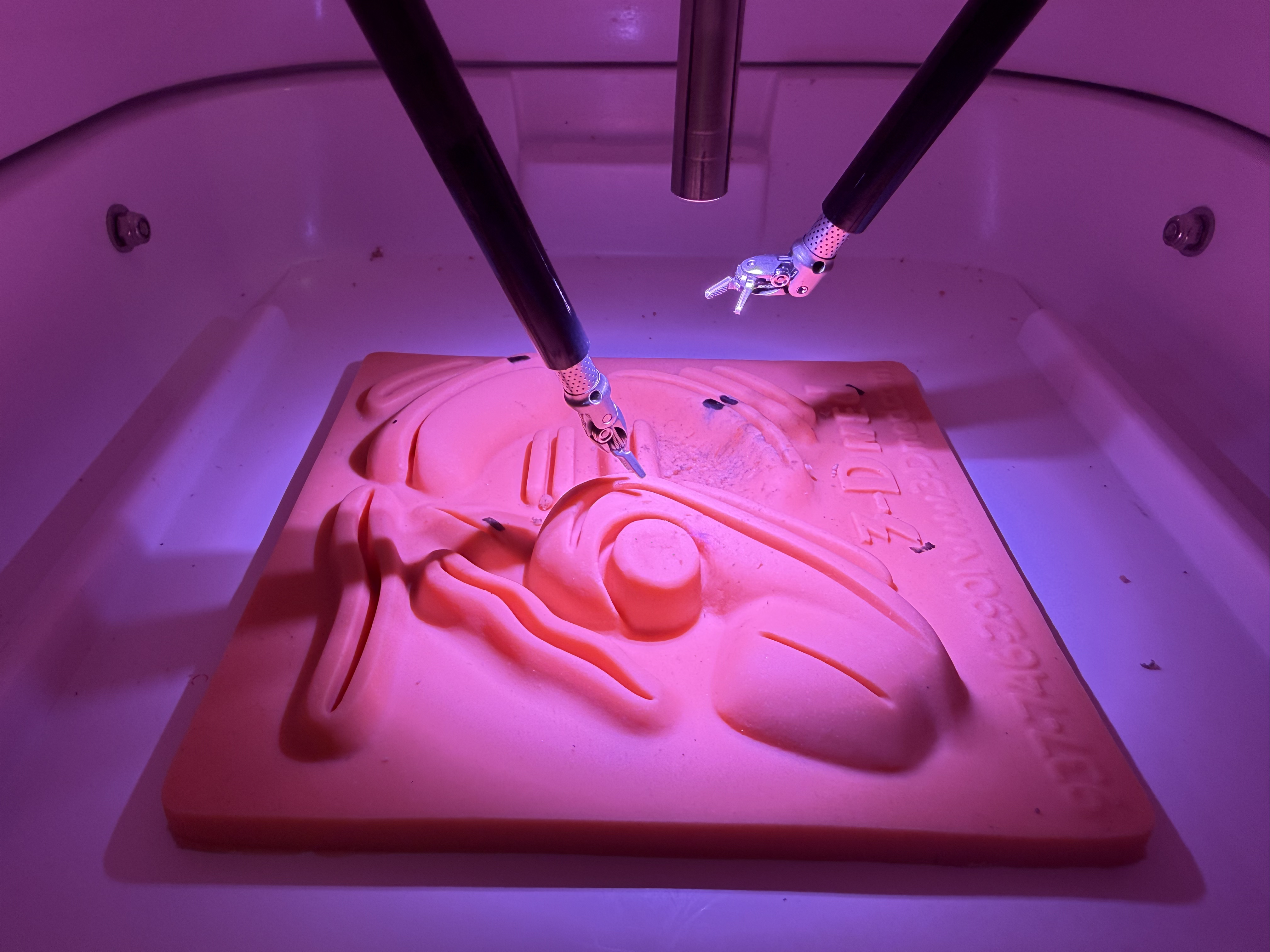}
    \end{subfigure}
    \hfill
    \begin{subfigure}[b]{0.48\linewidth}
        \centering
        \includegraphics[width=\linewidth]{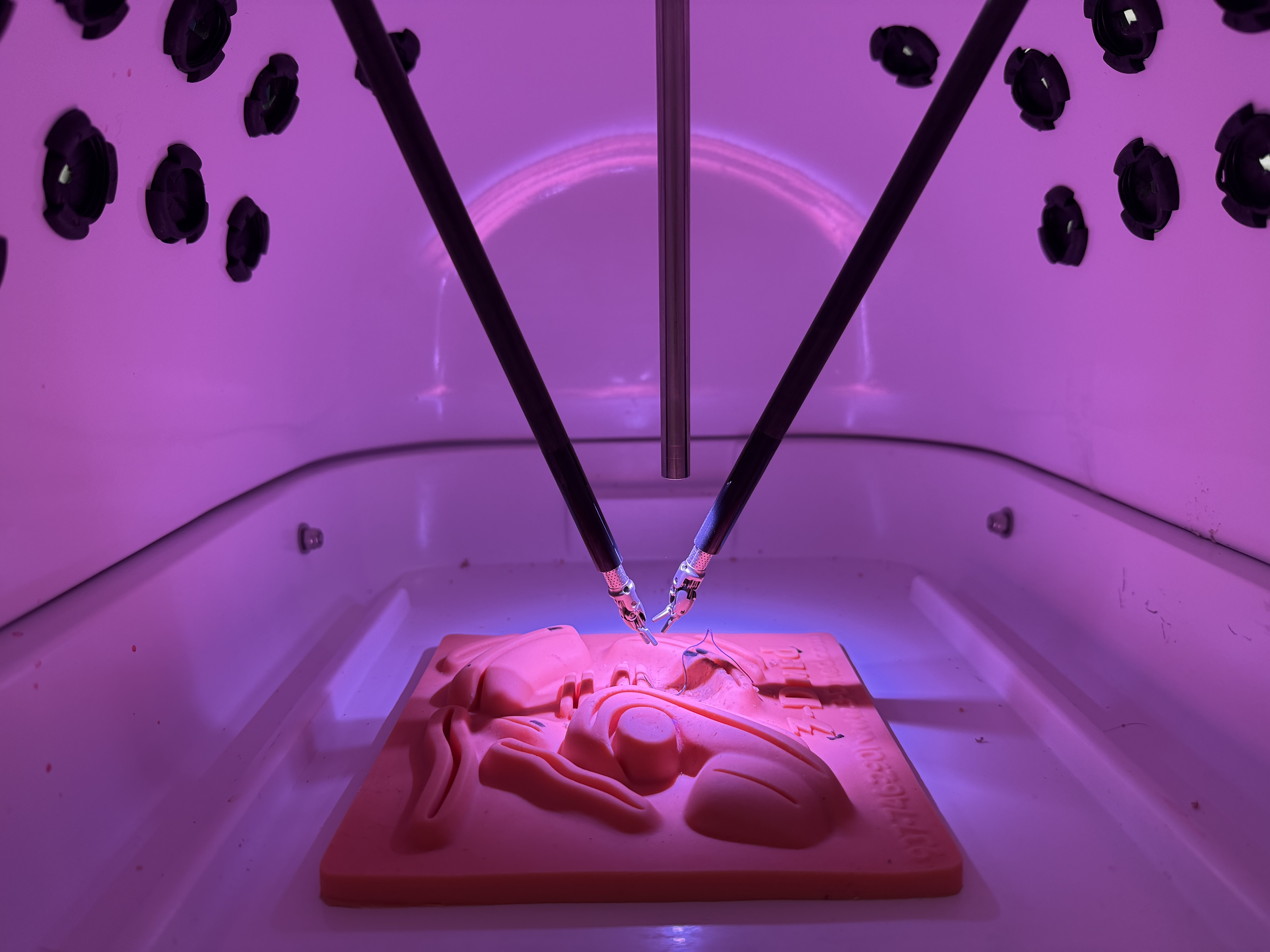}
    \end{subfigure}
    \caption{Physical setup for data collection and rollout for needle pickup and regrasp tasks. }
    \label{fig:supp_setup}
\end{figure}

We use a minimally invasive surgical robotics system equipped with a stereo endoscope and two bipolar forceps for gripping. For all kinematics, we operate in the end-effector Cartesian space from the camera space. We use relative translation deltas between each sample of each clip or episode and absolute rotations/gripper angles. This yields 13 values per arm (3 translation, 9 rotation, 1 gripper angle), or 26 values for two arms. 
Similar to prior work, we insert the endoscope and two arms into a mock abdominal cavity, where we also place a suture pad and a needle. For the sake of these experiments, we cut the needle thread short to improve needle pickup and regrasp; future work will use a needle with thread of proper length for the full suture. The physical set up of the robot can be seen in \Cref{fig:supp_setup}

\subsection{Tasks and needle configurations}
\label{sec:supp_tasks}
\paragraph{Needle pickup.} The needle pick up task is simply using the dominant arm (right) to grab and lift the needle. We first collect 557 demonstrations to train a baseline ACT model with. We rollout this ACT model and further collect 341 DAgger data to correct the policy when it starts to miss the needle. This yields 898 episodes. 3 trained operators performed data collection, and the average duration of the task is $4.9 \pm 3.4$\,s. The policy is deemed to be successful if it is able to pick up the needle within 3 tries and 500 model steps. 

\paragraph{Needle regrasp.} The needle regrasp task is comprised of 3 subtasks: (1) pick up the needle with the dominant arm, (2) hand it over to the non-dominant arm, and (3) hand it back to the dominant arm. This is an essential step before the dominant arm throws the suture to pierce through the tissue. We first collect 1889 demonstrations with one skilled operator. We train a baseline ACT model with these demonstrations and then collect 366 DAgger demonstrations to correct the policy with human intervention. In total, this yields 2255 episodes. The average duration of the task is $8.8 \pm 2.9$\,s. 

\begin{figure}[h]
  \centering
  \includegraphics[width=\linewidth]{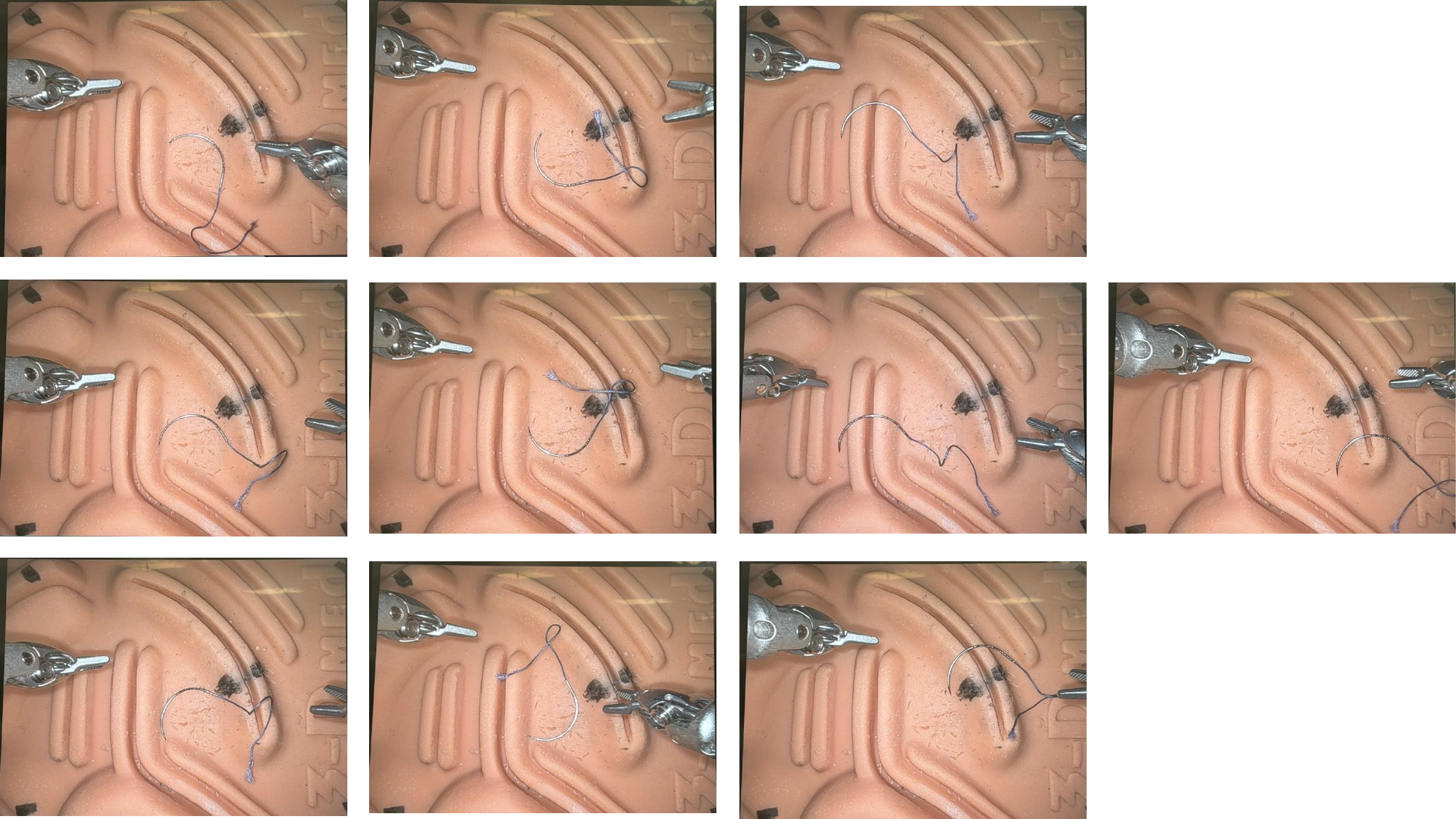}
  \caption{\textbf{Evaluation configurations.} The ten evaluation needle configurations for needle pickup and regrasp.}
  \label{fig:supp_configs}
\end{figure}

\paragraph{Evaluation configurations.} Every policy is evaluated on the same ten needle configurations, shown in \Cref{fig:supp_configs}. These are chosen for a wide variety of needle poses and positions throughout the suture pad. We initialize the arms at the beginning of each policy such that the grippers are along each side of the image boundaries. These needle configurations are \textit{outside} of the training set to ensure generalizability of the models. 
 
\section{Surgical DINOv2 Tokenizer}
\label{sec:supp_encoder}
\paragraph{Pretraining details.} We refer to the DINOv2 as a tokenizer rather than the encoder to reduce confusion with the ACT encoder. We pre-train the surgical vision model using the corpus described in Sec.~\ref{sec:data}. Training employs a dynamic video data loader that samples and decodes frames directly from surgical videos on the fly, rather than relying on a fixed set of pre-extracted frames. Following DINOv2~\cite{oquab2024dinov2}, we adopt a two-stage pretraining strategy. We first pretrain the ViT-L/14 at $224 \times 224$ resolution for 625K iterations, followed by 10K iterations of high-resolution adaptation at $518 \times 518$. This strategy enables the model to capture finer spatial details important for dense prediction tasks, such as segmentation and detection, without incurring the computational cost of high-resolution training from scratch. Both stages are trained on eight NVIDIA H200 GPUs, using reduced global batch sizes relative to the original DINOv2 configuration due to computational constraints. Additional training hyperparameters are provided in Tab.~\ref{tab:supp_encoder}.

\begin{table}[h]
  \centering\footnotesize
  \begin{tabular}{@{}lll@{}}
    \toprule
    Setting & Stage 1 & Stage 2 \\
    \midrule
    Architecture & ViT-L/14 & ViT-L/14 \\
    Input resolution & 224 $\times$ 224 & 518 $\times$ 518 \\
    Patch tokens & 256 & 1{,}369 \\
    Training iterations & 625K & 10K \\
    Global batch size & 1280 & 256 \\
    Optimizer & AdamW & AdamW \\
    Learning rate & 1.34e-4 & 2.5e-5 \\
    Training data & \multicolumn{2}{l}{Video frames (online sampling)} \\
    \bottomrule
  \end{tabular}
  \caption{\textbf{Surgical DINOv2 pretraining.}
  The model is trained in two stages using frames
  sampled dynamically from surgical videos.
  Since training uses an iterable dataset, the
  number of distinct frames encountered is not
  explicitly tracked.}
  \label{tab:supp_encoder}
\end{table}

\subsection{Vision model comparison study}

To ensure the use of the best vision model, we train vanilla ACT with (1) off-the-shelf DINOv2, trained on LVD-142M, (2) off-the-shelf DINOv3, trained on LVD 1689M, and (3) the surgical DINOv2 used in the paper. To ensure equal comparison with the in-house DINOv2, we use the ViT-L 4 register token variants, use IN-1k image normalization stats, and also train at 518px resolution. Since DINOv3 uses patch size of 16, we set the image size to $592$ instead so that the number of tokens are the same. 

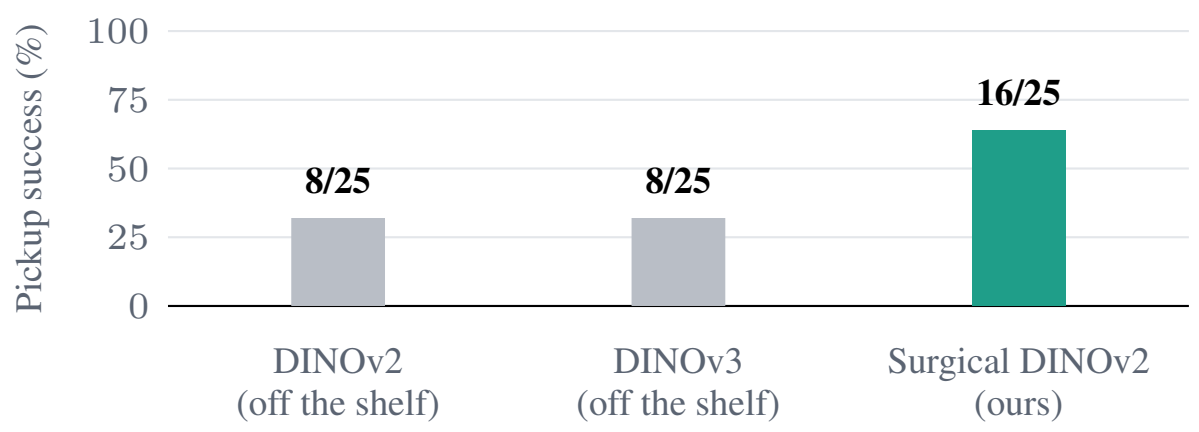
\begin{figure}[t]
  \centering
  \begin{tikzpicture}[font=\scriptsize]
    \begin{axis}[
        width=\linewidth, height=3.4cm, scale only axis=false,
        ybar, bar width=0.62cm,
        ymin=0, ymax=100, ytick={0,25,50,75,100},
        ylabel={Pickup success (\%)}, ylabel style={text=encMuted, font=\scriptsize},
        symbolic x coords={v2,v3,ih},
        xtick={v2,v3,ih}, bar shift=0pt,
        xticklabels={{DINOv2\\(off the shelf)}, {DINOv3\\(off the shelf)}, {Surgical DINOv2\\(ours)}},
        xticklabel style={align=center, text=encMuted, font=\scriptsize},
        enlarge x limits=0.25,
        axis x line*=bottom, axis y line*=left,
        y axis line style={draw=none},
        tick style={draw=none},
        yticklabel style={text=encMuted},
        ymajorgrids, grid style={encGrid, line width=0.4pt},
        nodes near coords,
        point meta=explicit symbolic,
        every node near coord/.append style={font=\scriptsize\bfseries, text=black, yshift=1pt},
      ]
      \addplot[fill=encOTS, draw=none] coordinates {(v2,32)[8/25] (v3,32)[8/25]};
      \addplot[fill=encIH, draw=none] coordinates {(ih,64)[16/25]};
    \end{axis}
  \end{tikzpicture}
  \caption{\textbf{Choice of vision model.} Needle-pickup success of ACT trained from scratch with each frozen DINO (25 trials: 5 configurations $\times$ 5 checkpoints, success within three attempts). The surgical DINOv2 doubles success over off-the-shelf DINOv2 and DINOv3.}
  \label{fig:supp_encoders}
\end{figure}

We train and evaluate on the needle pick up task over 5 needle configurations and 5 checkpoints. \Cref{fig:supp_encoders} shows the results. Off the shelf DINO models underperform compare to in-domain surgical models, despite the fact that suture pads are not in-vivo. However, surgical DINOv2 has been trained with images with surgical needles and robotic tools, which improves the downstream policy learning. 

\section{Pretraining and fine-tuning details}
\label{sec:supp_training}
 
\subsection{Architecture}
\label{sec:supp_arch}
The encoder input is a latent token, a robot-state token, and the 1{,}369 DINOv2 image tokens.
The decoder has 30 action queries and, during pretraining, 256 point queries.
Each point query is anchored at a pixel $u$: it is the sum of a projected Fourier embedding of $u$, a projection of the encoder output bilinearly sampled at $u$, and a learned query-type embedding.
Because the point features are read from the encoder output rather than from the frozen DINOv2 tokens, the flow loss trains the encoder both through these samples and through the decoder's cross-attention.
The flow head is a shallow MLP on the decoded point queries that predicts $P_0$ and the cumulative 3D displacement at each of the ten keyframes, trained with an L1 loss.

Point queries are appended to the action queries in a single decoder pass.
In all pretraining variants, a self-attention mask prevents the action queries from attending to the point queries, while the point queries may attend to the action queries.
The action queries therefore see the same inputs during pretraining as at fine-tuning, when the point queries and flow head are removed and the decoder runs on the action queries alone.
For flow-only pretraining, the robot state and the CVAE latent are set to zero and the action loss is off.
The action head then receives no gradient, but the action-query embeddings still do, because the point queries attend to them.

The architectural details of \method are shown in \cref{tab:supp_arch}.

\begin{table}[h]
  \centering\footnotesize
  \setlength{\tabcolsep}{4pt}
  \begin{tabular}{@{}lr@{}}
    \toprule
    \multicolumn{2}{@{}l}{\emph{Architecture}} \\
    Width / heads (head dim)             & 512 / 8 (64) \\
    Encoder / decoder layers             & 4 / 7 \\
    FFN width                            & 3{,}200 \\
    Image tokens                         & 1{,}369 (1024$\rightarrow$512-d) \\
    \midrule
    \multicolumn{2}{@{}l}{\emph{Parameters}} \\
    Transformer encoder                  & 17.33M \\
    Transformer decoder                  & 37.69M \\
    Input projection + positions         & 1.23M \\
    \textbf{Trunk (pretrained)}          & \textbf{56.25M} \\
    \addlinespace[2pt]
    Point queries + flow head (PT only)  & 0.56M \\
    State / latent tokens                & 0.03M \\
    Action queries + ACT action head     & 0.03M \\
    CVAE encoder (FT only)               & 17.41M \\
    \addlinespace[2pt]
    Total, excluding DINOv2              & 74.29M \\
    DINOv2 ViT-L/14 (frozen)             & 304.37M \\
    \bottomrule
  \end{tabular}
  \caption{\textbf{Policy architecture.} Parameter counts from the flow-only pretraining checkpoint (PT: pretraining, FT: fine-tuning). Only the trunk is pretrained and transferred to fine-tuning; the point queries and flow head are discarded, and the state and latent tokens, the CVAE encoder, and the action head start from their initialization. The flow-matching action head (\cref{sec:supp_fm}) adds 19.5M parameters in place of the CVAE encoder and the ACT action head.}
  \label{tab:supp_arch}
\end{table}

\subsection{Hyperparameters}\label{sec:supp_hparams}

Hyperparameters are shown in \Cref{tab:supp_hparams}.

\begin{table}[h]
  \centering\footnotesize
  \setlength{\tabcolsep}{3pt}
  \begin{tabular}{@{}lcc@{}}
    \toprule
    & Pretraining & Fine-tuning \\
    \midrule
    Data                    & clinical video & robot demos \\
    Iterations              & 100k                & 100k \\
    Batch size              & 512 & 128 \\
    Optimizer               & AdamW               & AdamW \\
    Learning rate (trunk)   & $8\times10^{-5}$    & $4\times10^{-5}$ \\
    LR schedule             & constant & constant$^\ast$ \\
    Weight decay            & $10^{-4}$           & $10^{-4}$ \\
    Precision               & bf16                & bf16$^\ast$ \\
    Action chunk            & --                  & 30 (2\,s) \\
    Point queries / anchor  & 256 ($\leq$128 movers) & -- \\
    Loss                    & L1 & ACT / FM \\
    Augmentation            & photometric & + state noise$^\dagger$ \\
    Checkpoint selection    & best val flow & every 20k steps \\
    GPUs / wall-clock       & 4 H200 & 1 A100 \\
    \bottomrule
  \end{tabular}
  \caption{\textbf{Training hyperparameters.} Both stages warm up the learning rate over 500 steps; it is $10^{-5}$ at a reference batch of 8, scaled by the square root of the batch size. Pretraining uses an L1 loss on $P_0$ and the displacements; fine-tuning uses ACT's L1 + KL ($\beta{=}10$) or the flow-matching loss. The DINOv2 is frozen and the input is the left image at $518{\times}518$ in both stages.
  $^\ast$With the flow-matching action head: cosine decay to 10\% of the peak rate, the head in fp32, and an EMA of the weights (decay 0.9995) for evaluation and checkpoints.
  $^\dagger$Gaussian noise on the robot state, 1.5\,mm and 3$^\circ$.}
  \label{tab:supp_hparams}
\end{table}
 
\section{Action heads and baselines}
\label{sec:supp_heads}
 
\subsection{Flow-matching action head}
\label{sec:supp_fm}
The FM head denoises the whole 30-step action chunk at once (19.5M parameters).
Each chunk step is a token: its noisy action is projected to the trunk width (512) and summed with a learned step embedding and a conditioning vector $c_\tau$, the decoder output of that step's action query plus a sinusoidal embedding of the flow time $t$.
Four transformer blocks (8 heads, MLP ratio 4) let the steps attend to each other, so a sampled chunk is one coherent trajectory, and $c_\tau$ modulates every attention and MLP sublayer through adaptive layer norm with zero-initialized gates (AdaLN-Zero~\cite{peebles2023dit}).

We use the linear (rectified-flow) path of flow matching~\cite{lipman2023flow}: for a normalized chunk $x_1$ and Gaussian noise $x_0$, $x_t = (1-t)\,x_0 + t\,x_1$ with $t \sim \mathcal{U}(0,1)$, and the head regresses the velocity $x_1 - x_0$ with a squared error, averaged over two $(t, x_0)$ draws per chunk and masked over padded steps.
At test time we integrate from noise with 10 Euler steps.
 
The head never sees the robot state directly.
The state enters, as in ACT, through the state token of the transformer encoder, which the action queries attend to in the decoder; the CVAE is not used with this head.
Actions are the same as for the ACT head (per arm: translation relative to the previous step, absolute rotation matrix, absolute gripper angle) and are z-scored per dimension with the training statistics of the initialization's data.
 
\subsection{Stabilizing flow-matching fine-tuning}
\label{sec:supp_fm_v2}
Our first FM runs (v1) diverged during fine-tuning: the training loss fell to about 0.07 and then rose to about 1.0 between 9k and 22k steps.
Three changes fixed this (v2): running the head in fp32 under mixed precision, a cosine learning-rate schedule decaying to 10\% of its peak, and an exponential moving average of the weights (decay 0.9995) used for evaluation.
All FM results in the paper use v2.
 
 
 
\subsection{Baselines}
\label{sec:supp_baselines}
\paragraph{Diffusion Policy.}
We train Diffusion Policy~\cite{chi2023diffusion} with its official implementation and its real-robot UNet recipe, replacing only the image tokenizer so that it matches \method: our frozen surgical DINOv2 at $518{\times}518$, whose tokens are pooled into 32 spatial-softmax keypoints and a 128-d global feature.
Demonstrations are converted with the same frame stride as ACT (every 4th native frame); the policy observes the left image and the 26-d robot state for $n_\text{obs}{=}2$ steps, predicts a horizon of 16 steps (64 native frames), and executes 8.
Actions use the same representation as ACT (per arm: translation relative to the previous step, absolute rotation matrix, absolute gripper angle), range-normalized as in Diffusion Policy.
The ConditionalUnet1D is trained with 100 DDPM noise levels (squared-cosine schedule, $\epsilon$-prediction) and sampled with 8 DDIM steps.
We train with AdamW (learning rate $10^{-4}$, 500 warmup steps, cosine decay), batch 64, and an EMA of the weights, for 60k gradient steps on one A100, and save six evenly spaced checkpoints.
Validation loss reaches its minimum near 20k steps and rises afterwards; we roll out 5 checkpoints similar to \method.
 
\paragraph{$\pi_{0.5}$ model.}
We fine-tune $\pi_{0.5}$~\cite{intelligence2025pi05} from the public base checkpoint with the \verb|openpi| code, both fully and with LoRA~\cite{hu2021lora}.
Each episode is paired with its task instruction, ``pick up the needle'' or ``pickup the needle and regrasp with the other tool''.
The model receives the left endoscope image and the 26-d robot state and predicts chunks of the same 26-d actions as ACT, normalized with statistics computed on the training demonstrations by openpi's normalization script.
LoRA fine-tuning uses 2 A100 GPUs and full fine-tuning 4; we roll out the checkpoints at 10k, 15k, 20k, 25k, and 30k steps.
 
\section{Full rollout results}
\label{sec:supp_rollouts}
\subsection{Main results}
\begin{figure*}[t]
  \centering
  \includegraphics[width=\textwidth]{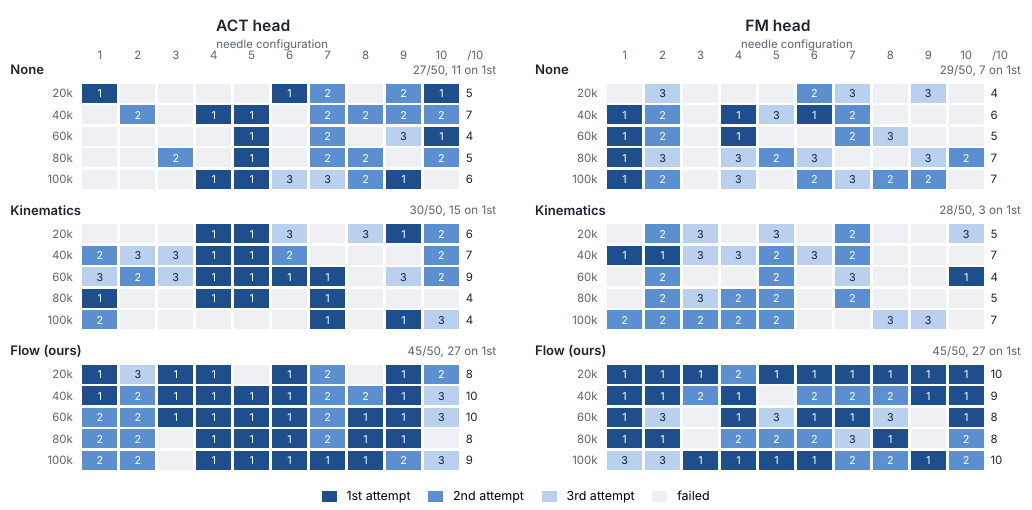}
  \caption{\textbf{Every needle-pickup trial.} Rows: pretraining and fine-tuning checkpoint; columns: the 10 needle configurations, the same for every policy. Colour and number give the attempt on which the needle was picked up; grey marks a failure within three attempts. Counts on the right are successes per checkpoint.}
  \label{fig:supp_pickup_rollouts}
\end{figure*}

\Cref{tab:main_results} reports success pooled over all checkpoints and configurations; here we break it down.
\Cref{tab:supp_pickup} gives successes per fine-tuning checkpoint, and \cref{fig:supp_pickup_rollouts} shows every trial.
The advantage of \method holds at every checkpoint and with both action heads, and it is not driven by a few configurations: the baselines fail repeatedly on the same configurations (2, 3, and 8) across checkpoints, where \method mostly succeeds, often on the first attempt.

\begin{table}[h]
  \centering\footnotesize
  \setlength{\tabcolsep}{3pt}
  \begin{tabular}{@{}llccccccc@{}}
    \toprule
    & & \multicolumn{5}{c}{Checkpoint} & \multicolumn{2}{c}{Total} \\
    \cmidrule(lr){3-7}\cmidrule(l){8-9}
    Pretraining & Head & 20k & 40k & 60k & 80k & 100k & 1st & $\le$3 \\
    \midrule
    None        & ACT & 5 & 7 & 4 & 5 & 6 & 11/50 & 27/50 \\
    None        & FM  & 4 & 6 & 5 & 7 & 7 &  7/50 & 29/50 \\
    Kinematics  & ACT & 6 & 7 & 9 & 4 & 4 & 15/50 & 30/50 \\
    Kinematics  & FM  & 5 & 7 & 4 & 5 & 7 &  3/50 & 28/50 \\
    Flow (ours) & ACT & 8 & 10 & 10 & 8 & 9 & 27/50 & \textbf{45/50} \\
    Flow (ours) & FM  & 10 & 9 & 8 & 8 & 10 & 27/50 & \textbf{45/50} \\
    \bottomrule
  \end{tabular}
  \caption{\textbf{Needle pickup, by fine-tuning checkpoint.} Successes within three attempts, out of the 10 needle configurations at each checkpoint, and totals over the 50 trials (first attempt and within three).}
  \label{tab:supp_pickup}
\end{table}

\begin{table}[h]
  \centering\footnotesize
  \setlength{\tabcolsep}{4pt}
  \begin{tabular}{@{}llccc@{}}
    \toprule
    Pretraining & Head & Pickup & Handover 1 & Full task \\
    \midrule
    None        & ACT & 28 & 4  & 2  \\
    Kinematics  & ACT & 46 & 12 & 2  \\
    Kin.\ + flow & ACT & 58 & 16 & 4 \\
    Flow (ours) & ACT & 76 & 52 & 14 \\
    \midrule
    None        & FM  & 36 & 20 & 0  \\
    Kinematics  & FM  & 14 & 4  & 0  \\
    Kin.\ + flow & FM & 30 & 6  & 2  \\
    Flow (ours) & FM  & 68 & 40 & 12 \\
    \midrule
    $\pi_{0.5}$ (LoRA) & -- & 20 & 10 & 0 \\
    $\pi_{0.5}$ (full) & -- & 24 & 8  & 0 \\
    Diffusion Policy   & -- & 42 & 18 & 2 \\
    \bottomrule
  \end{tabular}
  \caption{\textbf{Needle regrasp, all policies} (\% of trials completing each stage).}
  \label{tab:supp_regrasp}
\end{table}

\subsection{Data efficiency}
\label{sec:supp_rollouts_efficiency}

We fine-tune each initialization on a random 50\% of the demonstrations of each task, with the same architecture, schedule, and number of steps as at 100\%, and roll out the checkpoints at 20k, 40k, 60k, 80k, and 100k steps on the same 10 needle configurations (50 trials per policy).

\paragraph{Needle pickup.}
\Cref{tab:supp_eff_pickup} breaks down the 50\% pickup results.
With half of the demonstrations, flow pretraining nearly doubles success over training from scratch (52\% vs.\ 28\%) and over kinematics pretraining (26\%), and is the most consistent policy across checkpoints.
Kinematics pretraining does not improve on scratch.
For reference, scratch reaches 54\% with all demonstrations, so \method with half the data roughly matches scratch with all of it.

\begin{table}[h]
  \centering\footnotesize
  \setlength{\tabcolsep}{3.5pt}
  \begin{tabular}{@{}lcccccccc@{}}
    \toprule
    & \multicolumn{5}{c}{Checkpoint (k steps)} & \multicolumn{2}{c}{Total} & \\
    \cmidrule(lr){2-6}\cmidrule(lr){7-8}
    Pretraining & 20 & 40 & 60 & 80 & 100 & 1st & $\le$3 & vs.\ scratch \\
    \midrule
    None        & 1 & 2 & 6 & 2 & 3 &  6 & 14/50 (28\%) & -- \\
    Kinematics  & 0 & 3 & 5 & 4 & 1 &  7 & 13/50 (26\%) & 9 / 10 \\
    Flow (ours) & 6 & 3 & 4 & 7 & 6 & 11 & \textbf{26/50 (52\%)} & \textbf{22 / 10} \\
    \bottomrule
  \end{tabular}
  \caption{\textbf{Needle pickup with 50\% of the demonstrations} (ACT head). Checkpoint columns count successes within three attempts out of 10 configurations; ``1st'' counts successes on the first attempt. The last column gives paired wins / losses against scratch ($p=1.0$ and $p=0.050$).}
  \label{tab:supp_eff_pickup}
\end{table}

\paragraph{Needle regrasp.}
\Cref{tab:supp_eff_regrasp} reports the 50\% regrasp results next to the 100\% results.
\method fine-tuned on half of the demonstrations reaches more stages than scratch fine-tuned on all of them and than both other initializations at 50\%.
Halving the data costs \method about a third of its mean stages (1.42 to 0.98), but it remains the best policy at either data size.
Kinematics pretraining degrades sharply at 50\% (pickup 46\% to 16\%, no handovers), consistent with its reliance on the robot state (\cref{sec:supp_analysis}): with fewer demonstrations, the state shortcut is less constrained by visual evidence.
Scratch is near the floor at both data sizes (0.48 vs.\ 0.34 mean stages), so its 50\%-vs-100\% difference is not informative.

\begin{table}[h]
  \centering\footnotesize
  \setlength{\tabcolsep}{3pt}
  \begin{tabular}{@{}llcccccccc@{}}
    \toprule
    & & \multicolumn{5}{c}{Checkpoint (k steps)} & \multicolumn{3}{c}{Total (\%)} \\
    \cmidrule(lr){3-7}\cmidrule(l){8-10}
    Pretraining & Data & 20 & 40 & 60 & 80 & 100 & Pickup & HO1 & Full \\
    \midrule
    None        & 100\% & 1/0/0 & 4/0/0 & 1/0/0 & 3/1/1 & 5/1/0 & 28 & 4  & 2 \\
    None        & 50\%  & 1/0/0 & 2/0/0 & 4/1/0 & 5/2/1 & 3/3/2 & 30 & 12 & 6 \\
    \midrule
    Kinematics  & 100\% & 3/1/0 & 3/1/0 & 5/0/0 & 6/4/1 & 6/0/0 & 46 & 12 & 2 \\
    Kinematics  & 50\%  & 0/0/0 & 3/0/0 & 1/0/0 & 1/0/0 & 3/0/0 & 16 & 0  & 0 \\
    \midrule
    Flow (ours) & 100\% & 5/3/0 & 9/4/1 & 9/8/3 & 8/5/2 & 7/6/1 & \textbf{76} & \textbf{52} & \textbf{14} \\
    Flow (ours) & 50\%  & 5/2/0 & 6/3/2 & 5/3/0 & 5/3/0 & 8/5/2 & 58 & 32 & 8 \\
    \bottomrule
  \end{tabular}
  \caption{\textbf{Needle regrasp at 50\% and 100\% of the demonstrations} (ACT head). Checkpoint columns give pickup / handover 1 / full task out of 10 configurations; totals are over 50 trials. Mean stages reached (0--3): scratch 0.34 / 0.48, kinematics 0.60 / 0.16, flow 1.42 / 0.98 (100\% / 50\%).}
  \label{tab:supp_eff_regrasp}
\end{table}


\section{Additional analysis}
\label{sec:supp_analysis}
 
\begin{table}[h]
  \centering\footnotesize
  \setlength{\tabcolsep}{3pt}
  \begin{tabular}{@{}lcccc@{}}
    \toprule
    Pretraining target & Scratch & Kin. & Kin.\ + flow & Flow only \\
    \midrule
    \multicolumn{5}{@{}l}{\emph{Attention on the robot-state token (\%)}} \\
    Image tokens (PT)             & --   & 24.5 & 16.7 & 0.1$^\dagger$ \\
    Image tokens (FT)             & 7.7  & 21.3 & 14.7 & 7.9 \\
    Action queries (FT)           & 0.21 & 0.71 & 0.25 & 0.10 \\
    \midrule
    \multicolumn{5}{@{}l}{\emph{Offline error after fine-tuning}} \\
    Translation (mm / step)       & 0.31 & 0.30 & 0.31 & 0.30 \\
    Rotation (deg)                & 8.9  & 9.4  & 9.5  & 8.8 \\
    Jaw (L1)                      & 0.10 & 0.11 & 0.11 & 0.11 \\
    Perturbed $\Delta$ (mm)       & 0.20 & 0.19 & 0.30 & 0.19 \\
    \bottomrule
  \end{tabular}
  \caption{\textbf{Reliance on the robot state, and offline metrics.} Top: share of attention given to the robot-state token, from the image tokens in the encoder and from the action queries in the decoder, averaged over layers and heads, after pretraining (PT) or fine-tuning (FT). Bottom: action error on held-out demonstrations; perturbed $\Delta$ is the increase in translation error under 12 image perturbations (photometric changes, blur, noise, and occlusion). $^\dagger$The flow-only trunk is pretrained with a zero state, so its state token carries no information.}
  \label{tab:supp_state_offline}
\end{table}
 
\subsection{Reliance on the robot state}
\label{sec:supp_state}
\Cref{tab:supp_state_offline} (top) measures how much each policy attends to the robot-state token, averaged over layers and heads on held-out frames.
The kinematics-pretrained trunks route a large share of their attention to it.
In the encoder, the image tokens give about a quarter of their attention to the state token after kinematics pretraining (24.5\%) and about a sixth after kin.\ + flow pretraining (16.7\%).
This habit survives fine-tuning (21.3\% and 14.7\%), compared with about 8\% for both scratch and \method.
The action queries in the decoder show the same ordering at a much smaller scale, with the kinematics-pretrained policy attending most (0.71\%) and \method least (0.10\%).
 
\subsection{Offline metrics}
\label{sec:supp_offline}
The differences in \cref{sec:analysis} do not show up in the usual offline metrics (\cref{tab:supp_state_offline}, bottom).
On held-out demonstrations, the four fine-tuned policies have nearly the same action error: 0.30--0.31\,mm per step in translation, 8.8--9.5$^\circ$ in rotation, and 0.10--0.11 in jaw angle.
The offline test below, robustness to image perturbations, is similarly flat.

\paragraph{Image perturbations.}
We apply 12 perturbations to 400 held-out frames from the validation demonstrations: brightness scaled by 0.75 and by 1.25, contrast reduced to 0.6, saturation reduced to 0.4, gamma 0.7 and 1.4, a warm and a cool white-balance shift (red and blue channels scaled by 1.15 and 0.85 or the reverse), Gaussian blur with $\sigma = 2$ and 4\,px, additive Gaussian noise with standard deviation 0.03, and an occluding square covering 15\% of the image, filled with the image's mean color at a random position.
All perturbations act on the input image before the frozen DINOv2, so every policy receives the same perturbed features and differences come only from the trunk.
For each policy, we integrate the predicted chunk into per-arm 3D positions and measure how much farther they land from the demonstrated positions than on the clean frame, averaged over the 12 perturbations (perturbed $\Delta$).
The increase is small for every policy (0.19--0.30\,mm) and does not follow real-robot success: \method and training from scratch are equally robust by this measure.


\paragraph{Discussion.}
The policies differ only in closed loop, when their own actions determine what they see next, including states that the demonstrations rarely contain, such as recovering from a missed grasp.
The dense probes of needle geometry in \cref{sec:analysis} tracked real-robot success where offline action error did not, which echoes the finding that object-centric structure predicts policy success better than downstream accuracy~\cite{burns2023makes}.
We suggest probing what a policy encodes, not only how accurately it imitates, when comparing pretraining methods.

\subsection{Dense probe details}
\label{sec:supp_probes}
The dense probes of \cref{tab:probes} are linear: batch normalization followed by a $1{\times}1$ convolution on the $37{\times}37$ grid of trunk features at the image tokens, upsampled to the image resolution.
Semantic masks come from a segmentation model trained on internal data, and disparity from Fast-FoundationStereo~\cite{wen2026fast}. We use the needle pickup dataset demonstration data to perform these experiments. 

\begin{figure}
    \centering
    \includegraphics[width=\linewidth]{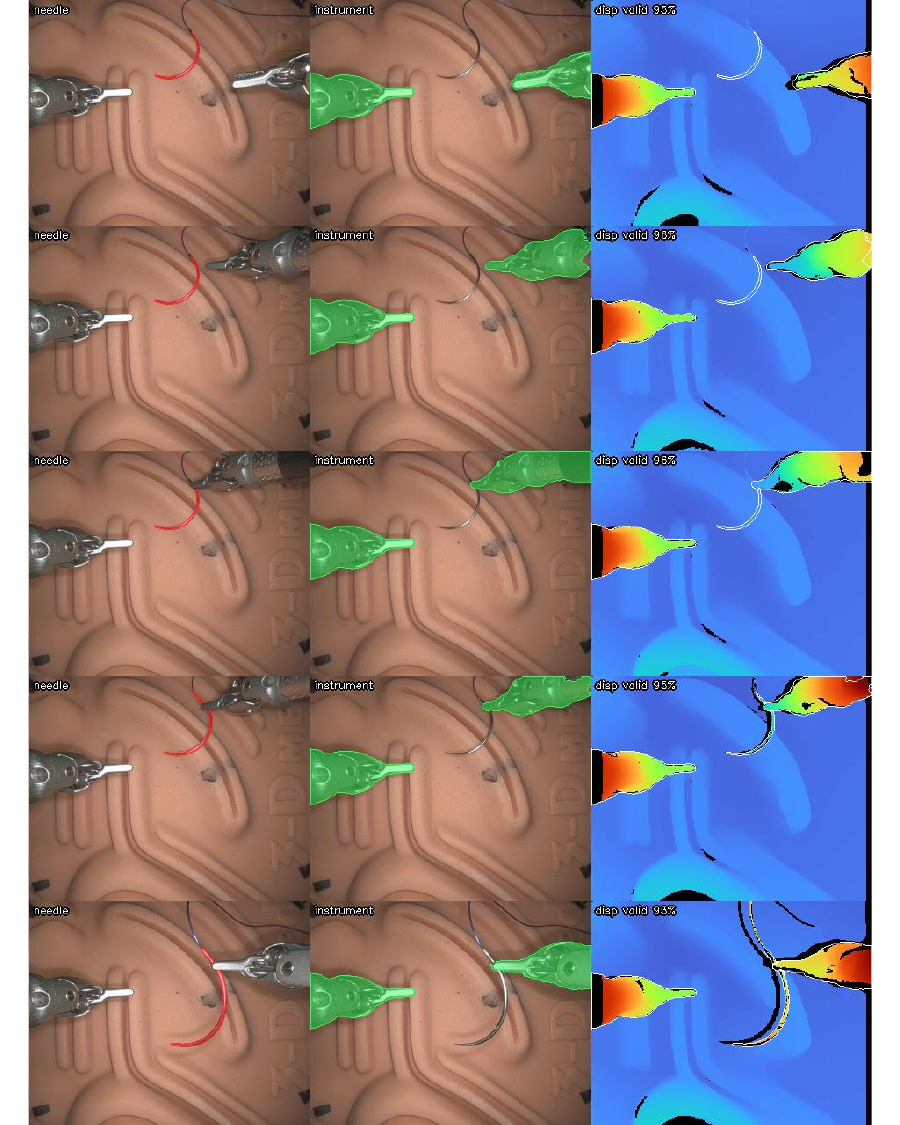}
    \caption{Samples of the needle segmentation and depth datasets. From left to right: RGB image with needle segmentation overlaid, RGB image with instrument segmentation overlaid, and metric depth map generated from Fast-FoundationStereo. Segmentation maps are generated with an internal needle segmentation model.}
    \label{fig:supp_probe_samples}
\end{figure}

To train the probes, we use the AdamW optimizer with a weight decay of \verb|1e-4| with a linear warmup and cosine decay scheduler. We sweep over learning rates $\{0.0003, 0.001, 0.003\}$ and show the highest performance. We use BCE DICE loss and train with 63152 frames and validate on 4539 frames. Examples of the labeled dataset for the probing experiments can be seen in \cref{fig:supp_probe_samples}. 

\subsection{Attention maps}
\label{sec:supp_attn}
\begin{figure}[t]
    \centering
    \includegraphics[width=\linewidth]{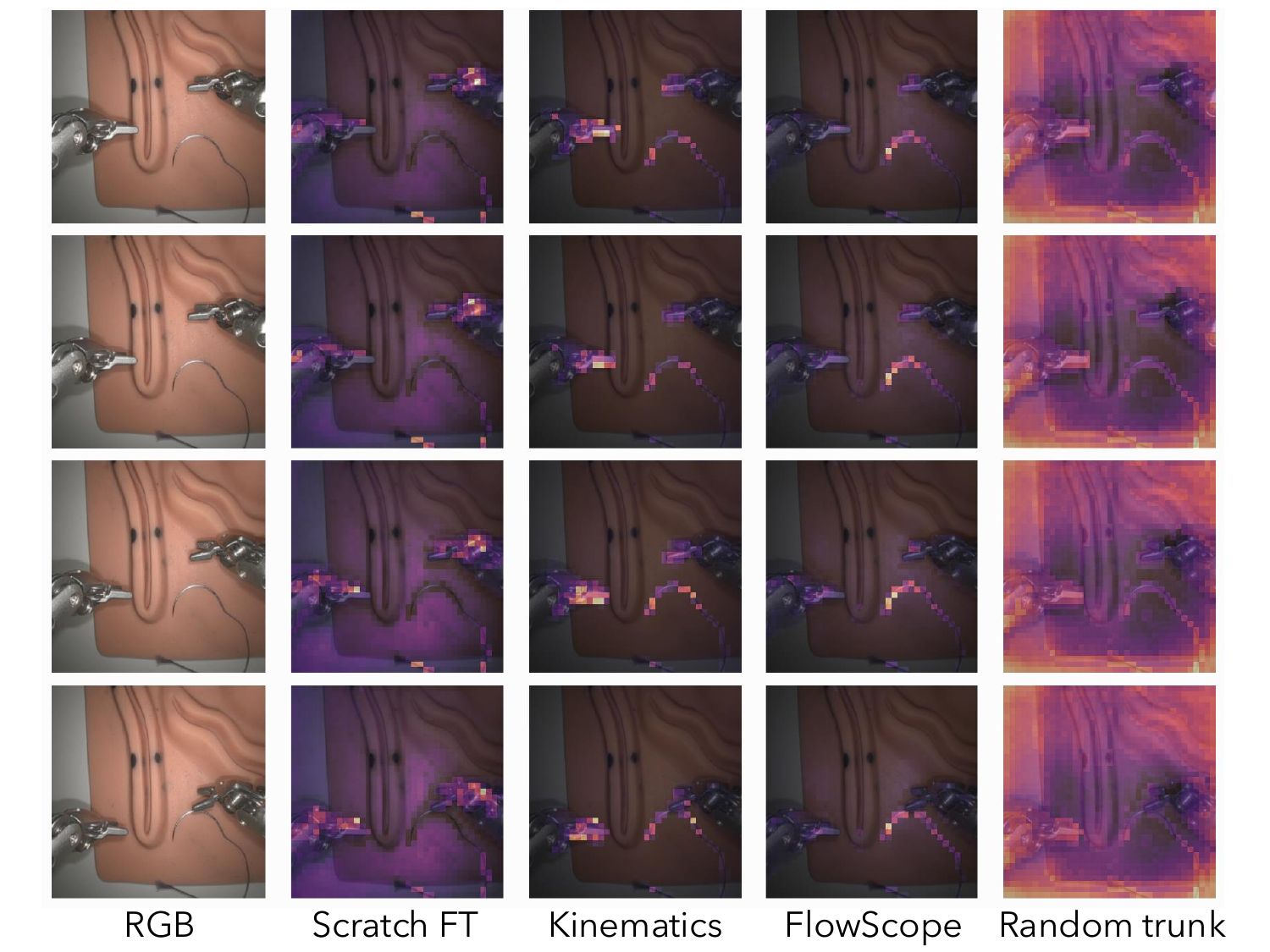}
    \caption{\textbf{Where the action queries look.} Decoder cross-attention from the action queries to the image, averaged over layers, heads, and queries, on held-out frames of needle pickup. Each map is normalized by its own maximum, limiting comparison between columns. From left to right: input image, training from scratch, kinematics pretraining, \method, and a randomly initialized policy (untrained action queries) for reference.}
    \label{fig:suppl_attnmaps}
\end{figure}

\Cref{fig:suppl_attnmaps} shows where the action queries of each fine-tuned needle-pickup policy look in the image, on held-out frames.
Each map is the decoder cross-attention from the 30 action queries to the image tokens, averaged over layers, heads, and queries, and normalized by its own maximum.
For reference, we also show a randomly initialized policy on the same frozen DINOv2 features, whose action queries are untrained.

The policy trained from scratch spreads its attention over the image with no clear focus on the needle or the instruments.
The kinematics-pretrained policy concentrates on the left instrument, which stays still during needle pickup.
The \method policy concentrates on the needle, with weaker peaks at the instrument tips.
These maps are qualitative and show where the policies read from, not what their actions depend on, but they agree with the probe results of \cref{tab:probes}: the policy whose trunk encodes the needle's depth best is also the one that looks at it.

\section{Additional experiments}
\label{sec:supp_exps}
 
 
 
\subsection{Scene flow as a co-training target}
\label{sec:supp_exp_cotrain}
Instead of pretraining on clinical video, scene flow can be predicted from the robot demonstrations themselves as an auxiliary target while the policy is trained.
We label the needle-pickup demonstrations with the same pipeline as the clinical video (Fast-FoundationStereo depth and SEA-RAFT flow chained over 10 keyframes, 10 frames apart, with the same consistency checks), using an anchor at every frame and storing 1{,}024 points per anchor, half uniform and half sampled in proportion to their motion.
During ACT training we add the point queries and flow head of \cref{sec:model} with the same settings as in pretraining (256 queries per sample, up to half movers above 5\,mm, L1 on $P_0$ and the displacements), weighted by 0.1 relative to the ACT loss, with action queries masked from the point queries so that the deployed policy is unchanged.
All other settings match the ACT baseline: frozen surgical DINOv2 at $518{\times}518$, chunks of 30 actions at stride 4, batch 128, 100k steps.

Co-training on the demonstrations alone raises needle-pickup success from 26/50 to 36/50 within the same rollout session (one seed).
This was our first sign that predicting scene flow shapes the trunk in a way that helps the policy, and it motivated pretraining the same head on clinical video, where far more motion is available than in the demonstrations.

\subsection{Stereo policies}
\label{sec:supp_exp_stereo}
\method uses stereo only to build its pretraining labels; the deployed policy sees the left image alone.
A natural alternative is to give the policy both eyes and let it extract geometry itself.
We test this with ACT, in the same setting as our other policies (frozen DINOv2 at $518{\times}518$, chunks of 30 actions at stride 4, batch 128, 100k steps, photometric augmentation with the same gains applied to both eyes). We build on StereoPolicy~\cite{han2026stereopolicy} for these experiments.

Both images pass through the same frozen DINOv2 and the same input projection.
The two token grids share their positional embeddings, since they cover the same pixel grid, and a learned eye embedding marks left and right.
We compare three variants.
\emph{Mono} uses the left eye only and is the matched control.
\emph{Concat} passes the left and right tokens ($2 \times 1{,}369$) directly to the ACT encoder.
\emph{Fusion} follows StereoPolicy~\cite{han2026stereopolicy}: before ACT, a two-layer stereo transformer (8 heads, feed-forward width 1{,}024, pre-norm) applies self-attention within each eye and then bidirectional cross-attention, the left tokens querying the right and vice versa, with weights shared between the eyes, so the tokens that reach ACT carry correspondence and disparity cues.
We run each variant with our surgical DINOv2 and with the off-the-shelf DINOv2.

\begin{table}[h]
  \centering\footnotesize
  \begin{tabular}{@{}llcc@{}}
    \toprule
    Encoder & Stereo input & Pickup 1st & Pickup $\le$3 \\
    \midrule
    Surgical DINOv2 & mono (left)   & 22 & \textbf{54} \\
    Surgical DINOv2 & concat        & \textbf{28} & 44 \\
    Surgical DINOv2 & fusion        & 20 & 52 \\
    \midrule
    DINOv2 (off the shelf) & mono (left) & 14 & 32 \\
    DINOv2 (off the shelf) & concat & 10 & 34 \\
    DINOv2 (off the shelf) & fusion & \textbf{18} & \textbf{40} \\
    \bottomrule
  \end{tabular}
  \caption{\textbf{Stereo at the policy input.} Needle-pickup success (\%) of ACT trained from scratch, with the left image or both eyes as input; $n=50$ trials per row, all rolled out in one session.}
  \label{tab:supp_stereo}
\end{table}

With our surgical DINOv2 tokenizer, neither stereo variant improves on the left image alone: fusion matches it (52\% vs.\ 54\%) and concatenation is lower (44\%), although concatenation succeeds slightly more often on the first attempt.
With the off-the-shelf DINOv2, fusion gives a modest gain over mono (40\% vs.\ 32\%), but all three variants remain well below the surgical encoder with a single view.
A domain-matched vision model thus matters more than a second view at the policy input.
Since stereo input also doubles the cost of the frozen ViT-L at every control step, we use the left view alone in all other experiments; \method instead exploits stereo where it is cheap and most useful, in constructing the pretraining targets.




 
 
\end{document}